# Recalibrating a Typed-Decision Model in an Afternoon: A Hands-On Recipe for Laya's Confidence Scores

*Measure, diagnose, fit, validate and deploy per-bucket temperatures, with numpy-only code and every number taken from a worked example*

Mohammed Nayeem Khan
mohammed.nayeem.khan@intel.com
mnknayeem@gmail.com

Aliyas Ahmad Misri
aliyas.ahmad.misri@intel.com

Intel Corporation

**Abstract**

A typed-decision model such as Laya reads a text and a list of options that the caller defines, for example ticket categories, and returns a probability for each option. Code around it compares the top probability with a threshold to answer, abstain or escalate to a person. Earlier work reports, for a Turkish model, that temperature scaling can raise held-out calibration error (ECE). This tutorial gives a five-step recipe (measure, diagnose, fit, validate, deploy) in 130 lines of numpy with standard methods, none new. On 1,065 synthetic, LLM-labelled items the released Laya checkpoint has a pooled ECE of 0.073 but a bucket-macro ECE of 0.143 (0.127-0.184), the mean over buckets (questions grouped by type and option count): the 12-option bucket is over-confident by +0.22, the three-to-five-option bucket under-confident by 0.05, and they cancel. One global temperature makes pooled ECE worse (0.173); a cross-fitted temperature per bucket brings macro-ECE to 0.073 (0.078 over 10 fold assignments) and reaches 0.089 with 10 labelled items per bucket, but holding out one of seven evaluation sets gives 0.099. After the refit, a gate aimed at 10% error answers 46.3% of items (10.5% wrong) against 6.9% on the published confidences, but one cut-off per bucket on the published confidences answers 45.4%, so most of the gain comes from separating buckets. Confidence-only maps (Platt, histogram, isotonic) need 20 or more labelled items per bucket to approach the temperature; with 320 per bucket the lowest macro-ECE belongs to Platt. In these files the retrained checkpoints store no per-bucket table and the stored 12-option value 0.1006 is clamped to 0.5 at runtime; the refit helps 31 of 33 checkpoints on random folds and 21 with a set held out. Fit and test share one generator and LLM labels, so these are findings about these checkpoints and this data.

**What you will learn (numbers recomputed from frozen results)**

1. Pooled ECE 0.073 hides bucket-macro ECE 0.143: choice:11+ is over-confident (+0.22) while choice:3-5 is under-confident (-0.05).
2. A single global temperature worsens pooled ECE (0.073 → 0.173); a per-bucket cross-fitted temperature gives macro-ECE 0.073.
3. The temperature is already near its floor with 10 items per bucket; Platt, histogram and isotonic maps need about 20, 40 and 80 items to get comparably low.
4. At a 10% target error a single cut-off answers 6.9% of items on published confidences versus 46.3% on refit confidences (errors 13.0% and 10.5%); one cut-off per bucket on published confidences answers 45.4% at 10.3%.
5. In the retrained checkpoints of this study none carries a per-bucket table (0 of 30); 25 of 30 store the choice temperature on the 5.0 ceiling, and the released config's 12-option value 0.1006 is clamped to 0.5.
6. The refit lowers macro-ECE for 31 of 33 checkpoints on random folds of this pool but for 21 of 33 when a whole set is held out; a map fitted on clean text recovers 63% of the benefit under the worst perturbation.

## 1 Why recalibrate, and what the recipe looks like

A typed-decision model is rarely used alone: somebody wraps it in a rule such as “answer if the confidence is at least 0.9, otherwise hand the ticket to a person”, or “send the item to the cheap path unless the confidence drops below a bar”. Those rules treat the confidence as a probability. If answers reported with a mean confidence of 0.95 are right only 72.2% of the time, as they are for one bucket of the model studied here, the rule fails silently: nothing crashes, but wrong answers arrive more often than the dashboard says.

This paper walks through the fix for someone who has the released Laya checkpoint (or one they trained) and an afternoon. It runs five steps on the released English checkpoint, evaluated on 1,065 synthetic support, hardware, sentiment and multilingual items, and prints every intermediate number. You can swap in your own data and reuse the code as printed: it is plain numpy, 130 non-blank lines in total, and every listing in the text is the exact source that produced the numbers beside it.

None of the methods here is new: temperature, Platt, histogram and isotonic calibration are standard, and earlier work on typed-decision models already benchmarks their probabilities (Section 2). The paper adds four things:

1. a measurement on the released Laya checkpoint, where pooled ECE hides a per-bucket error twice as large;
2. a worked recipe with ablations on how many labelled items each method needs, how fold seeds change the ranking of methods, and what a confidence gate delivers before and after;
3. a check on 33 checkpoints, all scored on this one pool, of where the per-bucket temperature helps;
4. facts about the temperature tables stored in the released and retrained checkpoints.

The checkpoint-level findings describe these files on this pool and say nothing about Laya’s trainer or other typed-decision models in general.

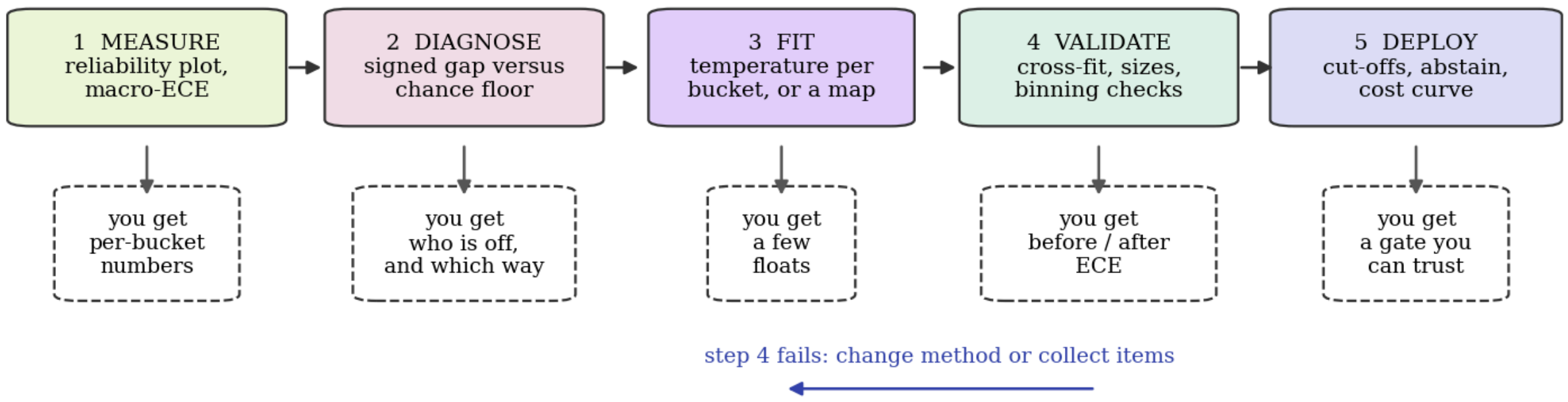

**Figure 1.** The five-step recipe. Each step produces a concrete artefact that the next step consumes; if step 4 fails, return to step 3 (another method) or collect more items. The numpy part runs in about 21 ms; labelling the calibration items is the real work.

You need a few hundred items from *your* traffic with a trusted label, ideally at least a few dozen per option-count bucket, and for each item the probability vector Laya returns (`probabilities`) or, better, the raw option logits. Beyond numpy, nothing else is required: no weights, GPU or retraining run. In code that is three aligned arrays per item: the probability vector, the index of the true option and the bucket. The numpy steps take about 21 ms for everything in this paper on 1,065 items; collecting labels is the part that takes the afternoon.

> **Conventions.** An *item* is one question answered once. *Bucket* is Laya’s own grouping of items by question type and number of options (for example `choice:11+` for questions with eleven or more options). *Confidence* is the top probability, `max(p)`. *ok* is 1 when the top option equals the label. All reliability numbers use ten equal-width confidence bins. Boxes headed “Check yourself” are small questions whose answers follow from the numbers just shown.

## 2 Related work and what is new here

Most work on the probabilities of typed-decision models is recent (Table 1). It covers calibration against a noise floor and the cost of decisions (TypedBench), temperature scaling failing on held-out questions (Teke), coherence (two papers, one on the English Laya checkpoint) and datasets with exact ground-truth probabilities. The classic calibration papers supply every method used below.

**Table 1.** Earlier work that this paper builds on or overlaps, with what each abstract reports and how this paper relates to it. Entries marked (preprint) are arXiv preprints.

| Work | What it reports | Relation to this paper |
|---|---|---|
| Guo et al. 2017 | Modern networks are poorly calibrated; temperature scaling, a one-parameter variant of Platt scaling, is surprisingly effective on most datasets tested. | First-line method here. The method is not ours; the per-bucket use on a typed-decision model is the measurement. |
| Platt 1999; Zadrozny and Elkan 2002; Niculescu-Mizil and Caruana 2005 | Sigmoid, histogram-binning and isotonic maps from a score to a probability. | Implemented in Listing 3 and compared with the temperature at several fit-set sizes (Step 4). |
| Kumar et al. 2019; Kull et al. 2019 | Platt and temperature scaling are less calibrated than reported, and histogram binning needs more samples (Kumar); Dirichlet calibration extends beta calibration to several classes (Kull). | Histogram binning does need more items here. We implement neither scaling-binning nor Dirichlet calibration, and we calibrate the top label only. |
| Nixon et al. 2019; Vaicenavicius et al. 2019; Błasiok et al. 2023 | ECE depends on binning choices, which can reorder recalibration methods; a framework for evaluating calibration; a kernel-smoothed ECE without bins. | Table 8 repeats the comparison under five binning rules. We do not compute smooth ECE or a debiased estimator. |
| Berta et al. 2026 (CalArena, preprint) | In about 2000 experiments on tabular and vision tasks, smooth calibration functions beat binning-based approaches. | Platt scaling overtaking the histogram and isotonic maps at 320 items per bucket points the same way, on one small pool. |
| Desai and Durrett 2020; Minderer et al. 2021 | BERT and RoBERTa are calibrated in-domain out of the box; recent non-convolutional image models are among the best calibrated. | Fits the small gaps of the retrained checkpoints (Section 13); the released checkpoint is the exception on this pool. |
| Sharma et al. 2026 (TypedBench, preprint) | Benchmark for typed decision models: calibration error against the finite-sample noise floor of a perfectly calibrated predictor, selective prediction, and realised cost under asymmetric costs. A hosted model is systematically underconfident. | The chance floor (Step 2), the gate and the cost curve (Step 5) apply the same ideas per bucket to Laya. They are not new here; the repair and its ablations are. |
| Teke 2026 (preprint) | For a Turkish typed-decision model, temperature scaling lowers smooth ECE on a development set but raises it on held-out support questions (0.027 to 0.045 for the first scored run). | A related warning. The failure seen here is different: one global temperature raises pooled ECE from 0.073 to 0.173, because the buckets need opposite corrections. Every ECE after refit is cross-fitted. |
| Li et al. 2026; Joy 2026 (preprints) | Probabilities of Jev miss the negation sum by 0.064 on average. On the English Laya checkpoint, probabilities rebuilt through broad categories differ from direct ones (mean total variation 0.424 to 0.689), and on MASSIVE ECE rose from 0.046 to 0.124 while accuracy rose 9.2 points. | Coherence is outside this paper, and we make no claim about it. |
| Rafe and Das 2026 (preprint) | Benchmarks decision models as automated gates; a guaranteed 5% risk leaves Jev 0.528 of the intent decisions. | Table 9 gives the same kind of coverage-at-target-error figures for Laya, per bucket. |
| Porcedda 2026; Dai et al. 2026 (preprints) | Datasets where the exact probability is known by construction (Sys1Cal-v1, ProbabilityBench), to test whether returned probabilities have the right meaning. | We have one observed label per item, so we test the top-label probability against outcomes, not the full distribution. |
| Santillana 2026; Sudjianto et al. 2026 (preprints) | Calibration of a perimeter-triage model collapses under temporal drift (ECE 0.04 to 0.40) and abstention thresholds transfer poorly across benchmarks; calibration corrects the scale of probabilities but not a biased teacher. | Agrees with the shift test (Section 12), where a clean-text map recovers part of the error. Our labels come from an LLM teacher (Limitations). |

Four phenomena in this paper were already known. ECE depends on the binning rule (Nixon et al.), a noise floor is needed to read ECE on small samples (TypedBench), temperature scaling can fail out of sample (Teke), and calibration degrades under shift (Santillana; Ovadia et al.). Steps 2, 4 and 5 re-measure them on Laya and none is presented as a finding. We found no earlier work that splits Laya's calibration by question type and option count, or that gives a recalibration recipe with a fit-set-size ablation for a typed-decision model. That search covered titles and abstracts in one index, so a closer match may exist.

## 3 What Laya's confidence numbers actually are

Before measuring anything, decide which number to calibrate. Laya answers every question of a state in one forward pass. Each option contributes one `[MASK]` marker, the encoder's hidden state at that marker is mapped by a small head to one logit, and the logits of a question are turned into probabilities. Figure 2 shows the part that matters here.

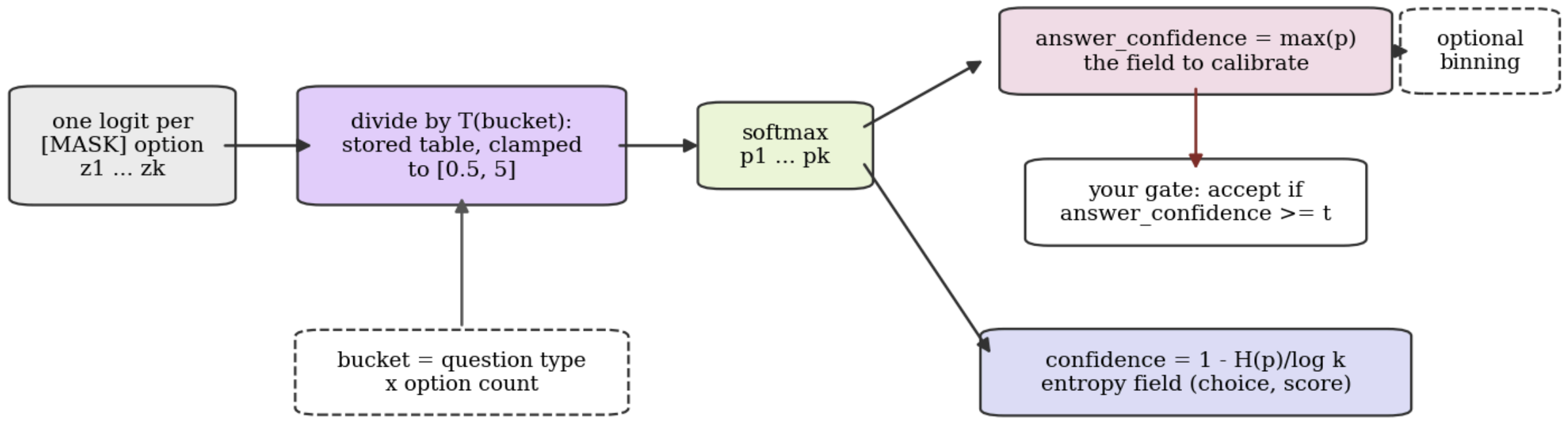

**Figure 2.** How Laya turns logits into the two confidence numbers it reports. The bucket decides which stored temperature divides the logits; the result feeds `answer_confidence` (probability of the reported answer) and, for choice and score questions, a separate entropy-based `confidence`.

First, the logits are divided by a temperature chosen by bucket, with the table stored in the checkpoint (Table 2); the runtime clamps every stored value into [0.5, 5]. A temperature below 1 sharpens the distribution, above 1 softens it. Second, the reply carries two different "confidences". `answer_confidence` is `max(p)`, the probability mass on the reported answer, and is the quantity temperature scaling is fitted for. The field called `confidence` is, for choice and score questions, a normalised entropy, `1 − H(p)/log k`, which measures how concentrated the whole distribution is; for two-option noul questions the two coincide.

**Table 2.** Temperatures stored in the released checkpoint's config, the value the runtime actually applies after its clamp to [0.5, 5], and how many items of the worked-example pool fall into each bucket.

| Bucket | Stored T | Applied T | Items in pool |
|---|---|---|---|
| choice:2 | 1.9064 | 1.9064 | 0 |
| choice:3-5 | 1.7602 | 1.7602 | 681 |
| choice:6-10 | 1.0000 | 1.0000 | 0 |
| choice:11+ | 0.1006 | 0.5000 (clamped) | 270 |
| score:3-5 | 1.2514 | 1.2514 | 89 |
| noul:2 | 1.9834 | 1.9834 | 25 |

The two fields correlate strongly (0.95 on the 1040 choice and score items) but sit at different levels. Accuracy on these items is 68.8%; the mean of `max(p)` is 0.718, so it is roughly the right size, while the mean entropy field is only 0.469. Read as a probability, the entropy field has ECE 0.307, against 0.072 for `max(p)`. As a *ranking* of which items to trust, the entropy field is almost as good (area under the risk-coverage curve 0.197 versus 0.187; lower is better), so the two differ mainly in calibration. A threshold set on one field does not carry over to the other, and the entropy field drifts with the number of options (Figure 3, right), so everything below calibrates `answer_confidence`.

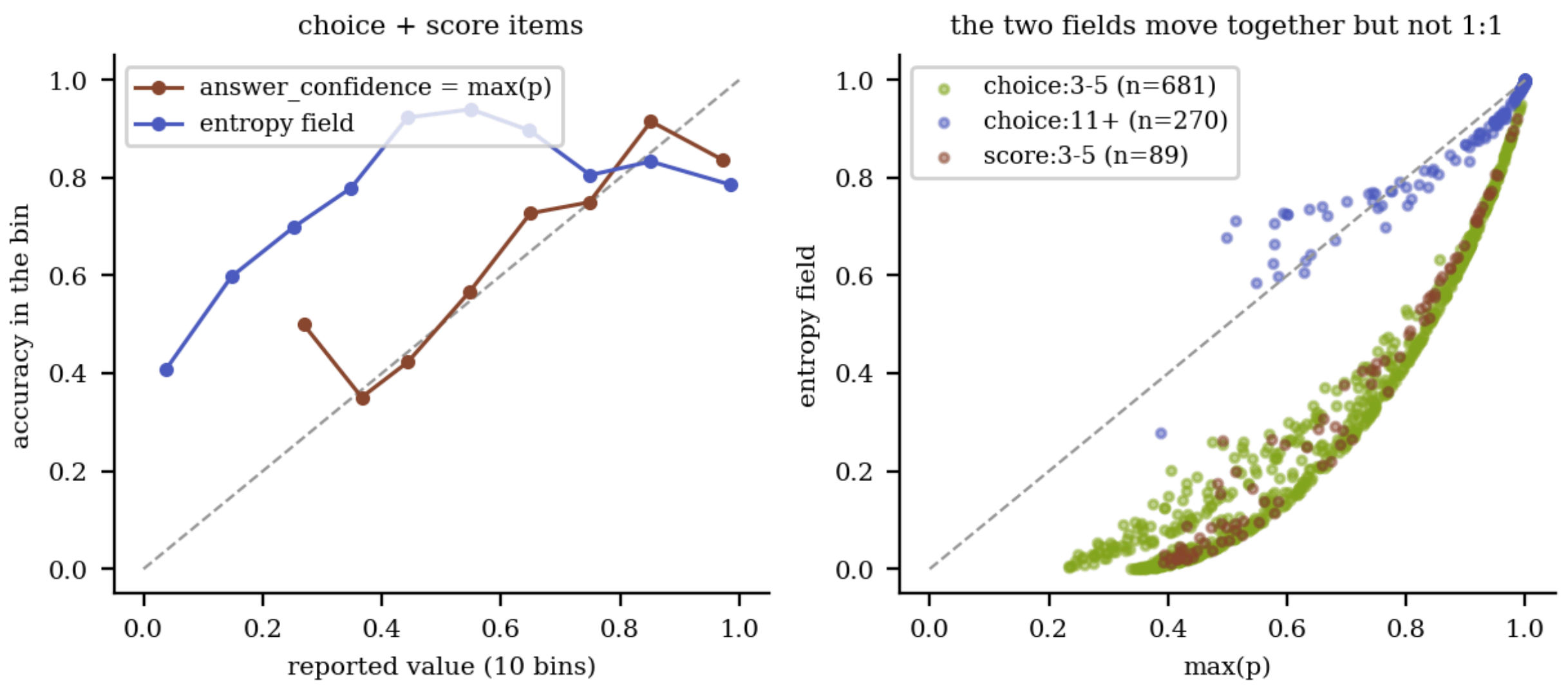

**Figure 3.** Left: reliability of the two confidence fields on choice and score items (a calibrated field would follow the dashed diagonal). Right: the entropy field against max(p), coloured by bucket; the relation depends on the number of options.

**Table 3.** Mean value of each confidence field against accuracy, per bucket (choice and score items; the noul field is max(p) by definition).

| Bucket | n | Accuracy | Mean max(p) | Mean entropy field |
|---|---|---|---|---|
| choice:3-5 | 681 | 68.7% | 0.633 | 0.299 |
| choice:11+ | 270 | 72.2% | 0.947 | 0.943 |
| score:3-5 | 89 | 58.4% | 0.667 | 0.336 |

**Check yourself.** In Table 3, which bucket has a mean `max(p)` far above its accuracy? (choice:11+: 0.947 against 0.722.) The entropy field there is 0.943, almost the same as `max(p)`, because with twelve options a distribution has to be very peaked to look confident at all. Remember this bucket; it returns in every step.

## 4 The worked example: model, data and caveats

The model. Laya (*convaiinnovations/laya*) is a ModernBERT-large encoder [Warner et al. 2024] with a two-layer option head and 421 million parameters in the released file. It is non-autoregressive: one forward pass reads a hidden state at the marker of every option and returns a typed answer (a choice, a score or a yes/no) for each question. We call the released English checkpoint *base* throughout; Section 13 repeats the recipe on 33 checkpoints, this one included.

The data. We pooled 1,065 items from seven evaluation sets (Table 4). Each item was written by an LLM (Qwen3.8-27B) conditioned on a target label, then re-labelled blind by a second pass of the same model; items on which the two passes disagreed were dropped. An independent model (gemma4-31b) re-labelled the final items and agrees with the stored label on 94%-99% of them depending on the set. All evaluation sets were de-duplicated against the 2,818-row training set used for the fine-tuned models in Section 13, though near-duplicates from the same generator cannot be excluded.

**Table 4.** The seven evaluation sets pooled for the worked example: size, accuracy and mean confidence of the released checkpoint, and which buckets the items fall into.

| Set | Content | n | Acc. | Mean conf. | ECE | Bucket mix (items) |
|---|---|---|---|---|---|---|
| probe | mixed probe set, nine categories | 224 | 71.9% | 0.759 | 0.133 | ch 3-5 113, ch 11+ 75, sc 3-5 11, nl 2 25 |
| sent_test | sentiment, clean text | 106 | 93.4% | 0.864 | 0.074 | ch 3-5 106 |
| json_test | tickets serialised as JSON | 58 | 74.1% | 0.697 | 0.183 | ch 3-5 58 |
| hw_test | hardware failure triage | 82 | 62.2% | 0.455 | 0.169 | ch 3-5 82 |
| many_test | 12-way support topic | 195 | 73.3% | 0.947 | 0.214 | ch 11+ 195 |
| ml_test | 12 languages, English-trained model | 322 | 61.2% | 0.595 | 0.058 | ch 3-5 322 |
| urg_test | urgency score | 78 | 57.7% | 0.666 | 0.143 | sc 3-5 78 |

Three caveats apply to every number in the tutorial. The labels are LLM-made, so “accuracy” means agreement with the teacher, and a calibration fitted here would not transfer to a deployment unless the labels there are of similar kind. Every item is synthetic. Simple baselines are strong on this data: a TF-IDF plus logistic-regression classifier trained on the same rows reaches 88% on hw_test, 91% on urg_test, 84% on json_test, 88% on many_test, and it matches or beats the fine-tuned Maya-v1 checkpoint on 1 of 5 sets (ml_test). A calibrated Laya is worth deploying only if it clears that accuracy bar on your task. The baseline's own probabilities could be recalibrated with the same recipe; we did not do that.

## 5 Step 1: measure with reliability diagrams, and the pooling trap

The first job is to see the problem: group the items by the confidence they were answered with, and compare each group's mean confidence with its accuracy. If the model is calibrated the two agree in every group. Listing 1 is all the code needed. `ece` is the count-weighted average of the per-bin gaps [Naeini et al. 2015], the numbers behind a reliability diagram [DeGroot and Fienberg 1983], and `macro_ece` averages that quantity over buckets so that each bucket counts once.

```
import numpy as np

def bin_index(conf, bins=10):
    """Equal-width bins (lo, hi]; confidence 1.0 lands in the last one."""
    return np.clip(np.ceil(np.asarray(conf, float) * bins).astype(int) - 1, 0, bins - 1)

def reliability(conf, ok, bins=10):
    """Rows of (mean confidence, accuracy, count): the numbers behind a reliability diagram."""
    conf, ok = np.asarray(conf, float), np.asarray(ok, float)
    idx = bin_index(conf, bins)
    return [(conf[idx == b].mean(), ok[idx == b].mean(), int((idx == b).sum()))
            for b in range(bins) if (idx == b).any()]

def ece(conf, ok, bins=10):
    """Expected calibration error: count-weighted |confidence - accuracy| over the bins."""
    return sum(c * abs(m - a) for m, a, c in reliability(conf, ok, bins)) / len(conf)

def macro_ece(conf, ok, bucket, bins=10):
    """Mean of the per-bucket ECEs, so a 25-item bucket counts as much as a 700-item one."""
    conf, ok, bucket = np.asarray(conf), np.asarray(ok), np.asarray(bucket)
    return float(np.mean([ece(conf[bucket == u], ok[bucket == u], bins) for u in np.unique(bucket)]))
```

**Listing 1.** Measuring calibration. The bins are (lo, hi], so a confidence of exactly 0 falls into the first bin and exactly 1 into the last; this matters later, because some recalibration maps output exactly 0 or 1.

Run on the pooled 1,065 items this gives ECE 0.073 (95% bootstrap interval 0.062-0.103), a value that looks acceptable on its own. Run it again *per bucket* and average, and the macro-ECE is 0.143 (0.127-0.184), about 2.0 times larger. In Figure 4 the pooled diagram sits close to the diagonal, but the individual buckets do not: choice:3-5 is under-confident by 0.05, choice:11+ is over-confident by 0.22. The remaining buckets (score:3-5, noul:2) are too small to tell, as Step 2 will show. A bucket that is too sharp and a bucket that is too flat cancel in the pooled average, and the larger bucket (here choice:3-5, 681 items) dominates it.

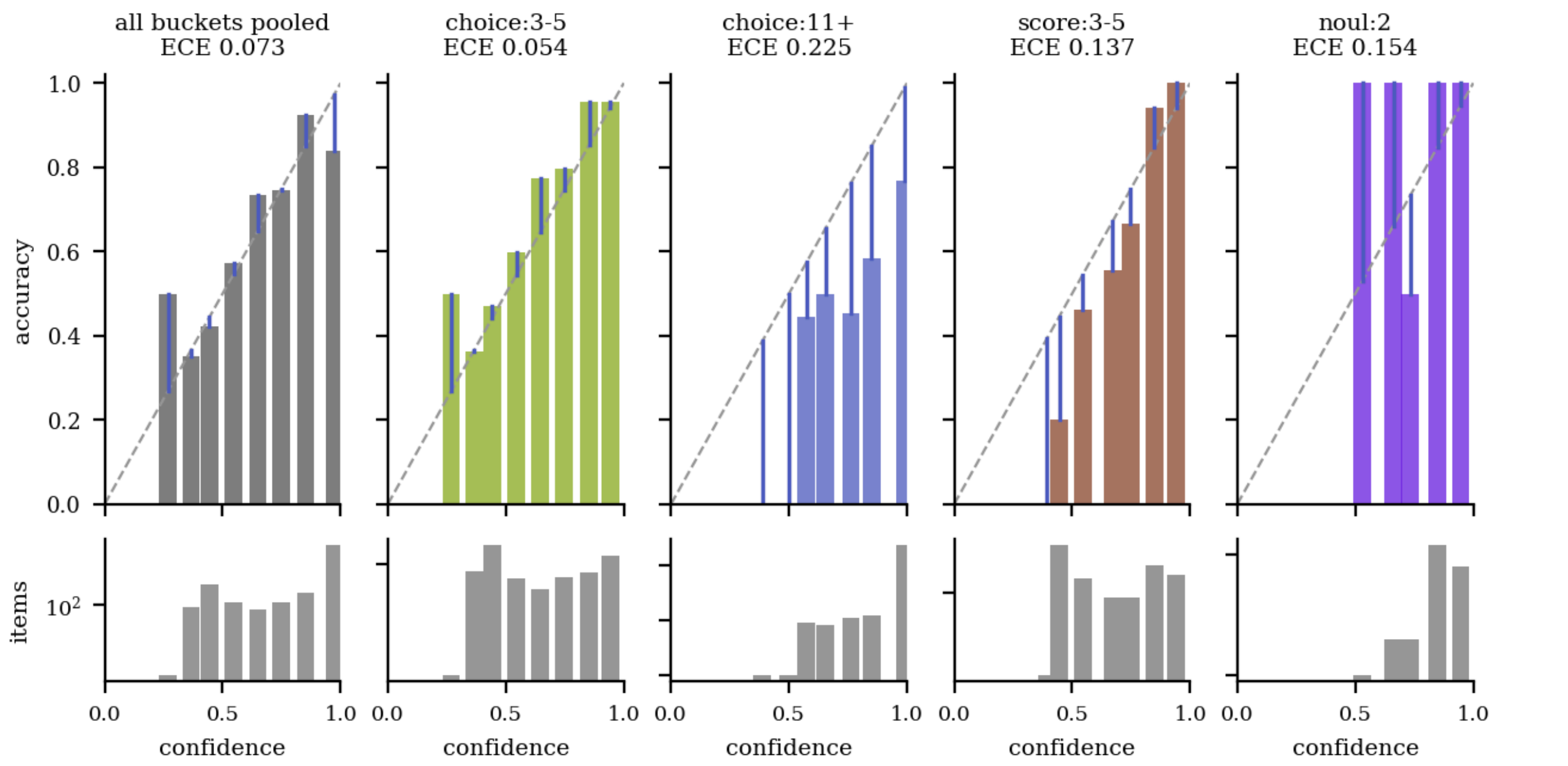

**Figure 4.** Reliability diagrams of the base checkpoint, pooled (left) and per bucket. Bars: accuracy in each confidence bin; red ticks: the gap to the diagonal; lower row: number of items per bin on a log axis. The pooled panel looks tame because the per-bucket gaps point in different directions.

## 6 Step 2: diagnose who is over- and who is under-confident

Measuring told you *that* something is off; diagnosing tells you *where*, and whether the effect is real or just noise from a small bucket, in three checks.

### 6.1 Compare each bucket with the chance floor

ECE is biased upwards in small samples: even a perfectly calibrated confidence scores above zero because the observed accuracy in each bin is a noisy estimate. Figure 5 simulates that floor, the same idea TypedBench [Sharma et al. 2026] applies to typed-decision models. For the large buckets it is low (about 0.05 for the 681 items of choice:3-5); for the 25-item `noul:2` bucket it is 0.20. Table 5 marks a bucket as over- or under-confident only if its ECE is above its own floor; 2 of 4 buckets are within chance (score:3-5, noul:2), 1 is over-confident (choice:11+) and 1 under-confident (choice:3-5).

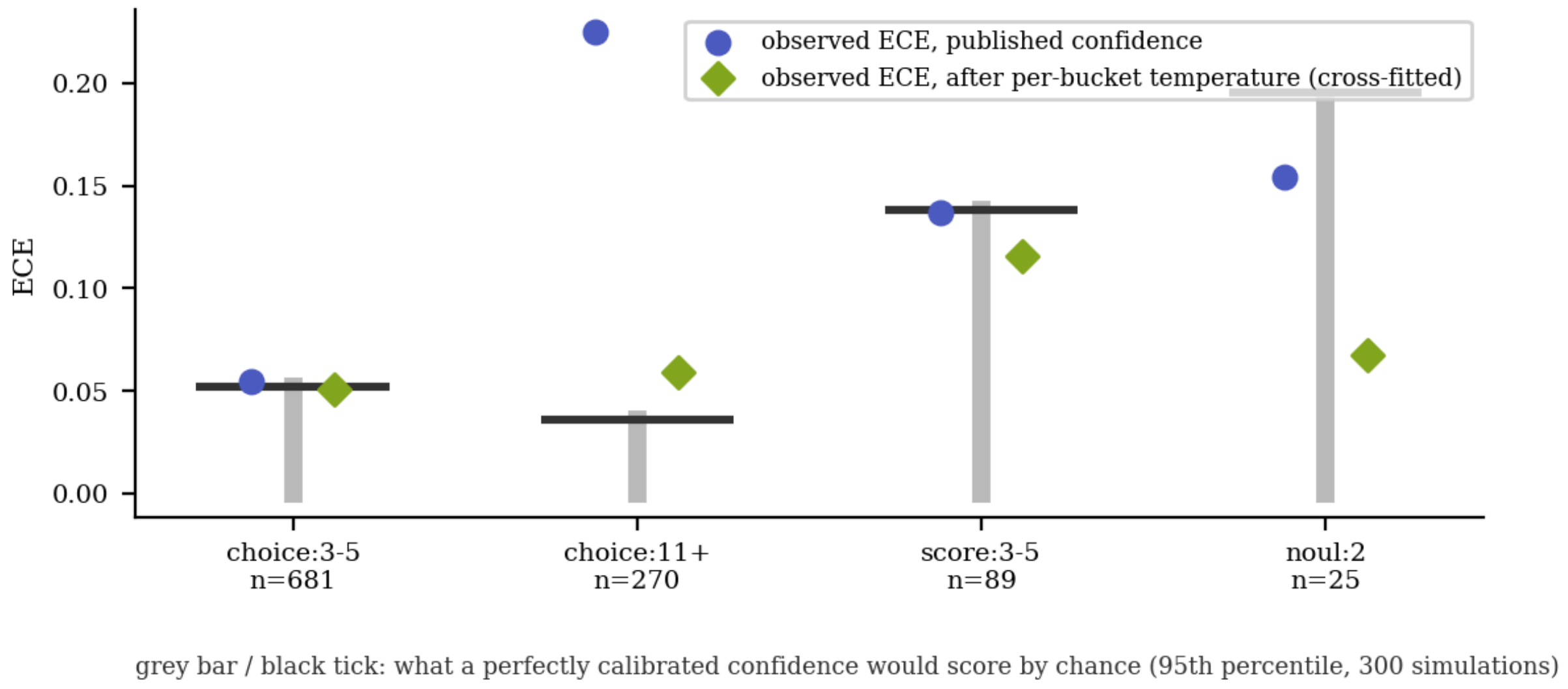

**Figure 5.** How large an ECE has to be before it means something. For each bucket the grey bar shows the ECE a *perfectly calibrated* model would still score because of finite samples (95th percentile of simulations that draw each item's correctness from its own confidence). Small buckets have a high floor.

**Table 5.** Per-bucket diagnosis of the base checkpoint. Gap = mean confidence minus accuracy (positive: over-confident). "Chance ECE" is the 95th percentile of the ECE of a perfectly calibrated confidence with the same n and confidences.

| Bucket | n | Accuracy | Mean conf. | Gap | ECE | Chance ECE (p95) | Reading |
|---|---|---|---|---|---|---|---|

| Bucket | n | Accuracy | Mean conf. | Gap | ECE | Chance ECE (p95) | Reading |
|---|---|---|---|---|---|---|---|
| choice:3-5 | 681 | 68.7% | 0.633 | -0.054 | 0.054 | 0.052 | under-confident |
| choice:11+ | 270 | 72.2% | 0.947 | +0.225 | 0.225 | 0.036 | over-confident |
| score:3-5 | 89 | 58.4% | 0.667 | +0.083 | 0.137 | 0.138 | within chance |
| noul:2 | 25 | 96.0% | 0.843 | -0.117 | 0.154 | 0.195 | within chance |

## 6.2 Look at the signed gap, not the absolute one

ECE hides direction, but a gap with a sign tells you which way to turn the temperature: over-confident (confidence above accuracy) calls for a *higher* temperature, under-confident for a *lower* one. Multiplying each bucket's gap by its share of the items and stacking the results (Figure 6) reproduces the pooled mean gap of +0.027, a number small enough to ignore, from contributions as large as +0.057.

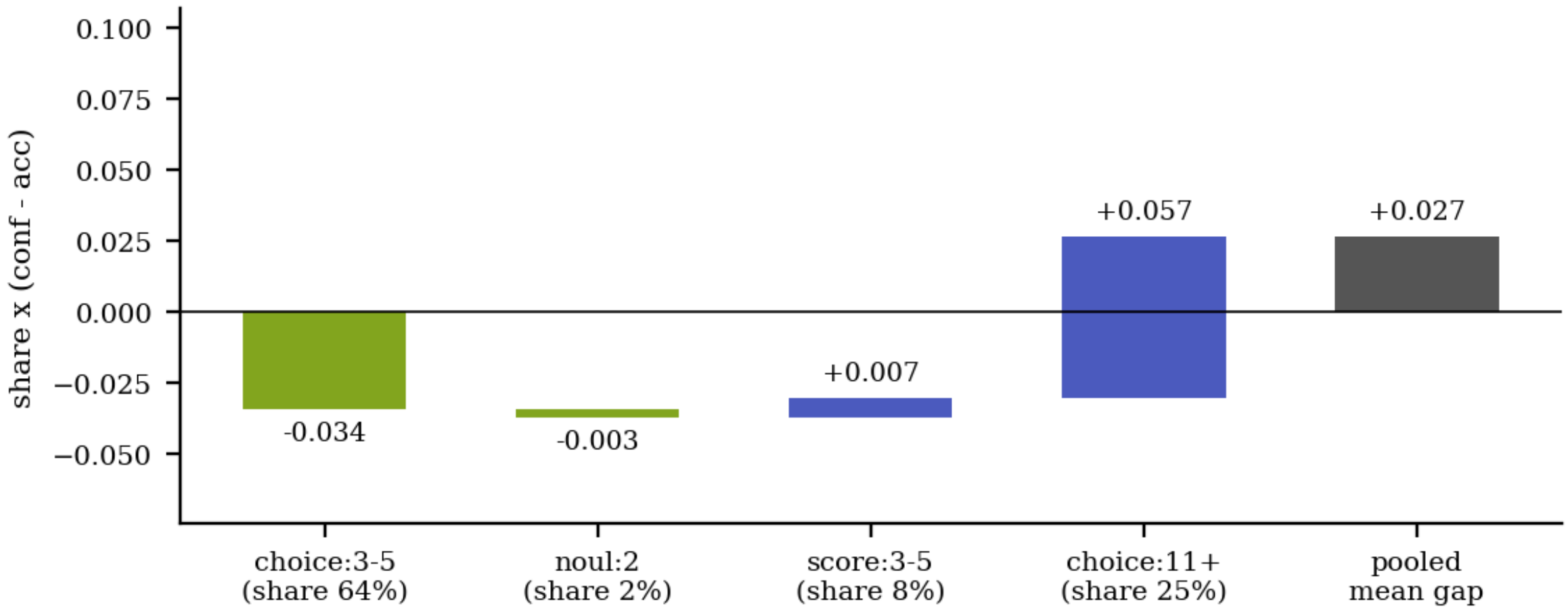

**Figure 6.** Waterfall of the signed gap. Each bar is a bucket's share of the items times its confidence-minus-accuracy gap; blue (under-confident) and red (over-confident) contributions add up to the small grey bar, which is all that a pooled diagnostic sees.

To see *where* in the confidence range the gap lives, split each bucket by confidence (Figure 7). The over-confidence of choice:11+ sits in the 0.6-0.8 confidence range, where its gap is +0.24 over 19 items, and 240 of its 270 items are reported above 0.8. Because most of the bucket sits above 0.8, a cut-off such as 0.9 does not avoid it: the 0.8-1.0 bin has a gap of +0.23 over 240 items.

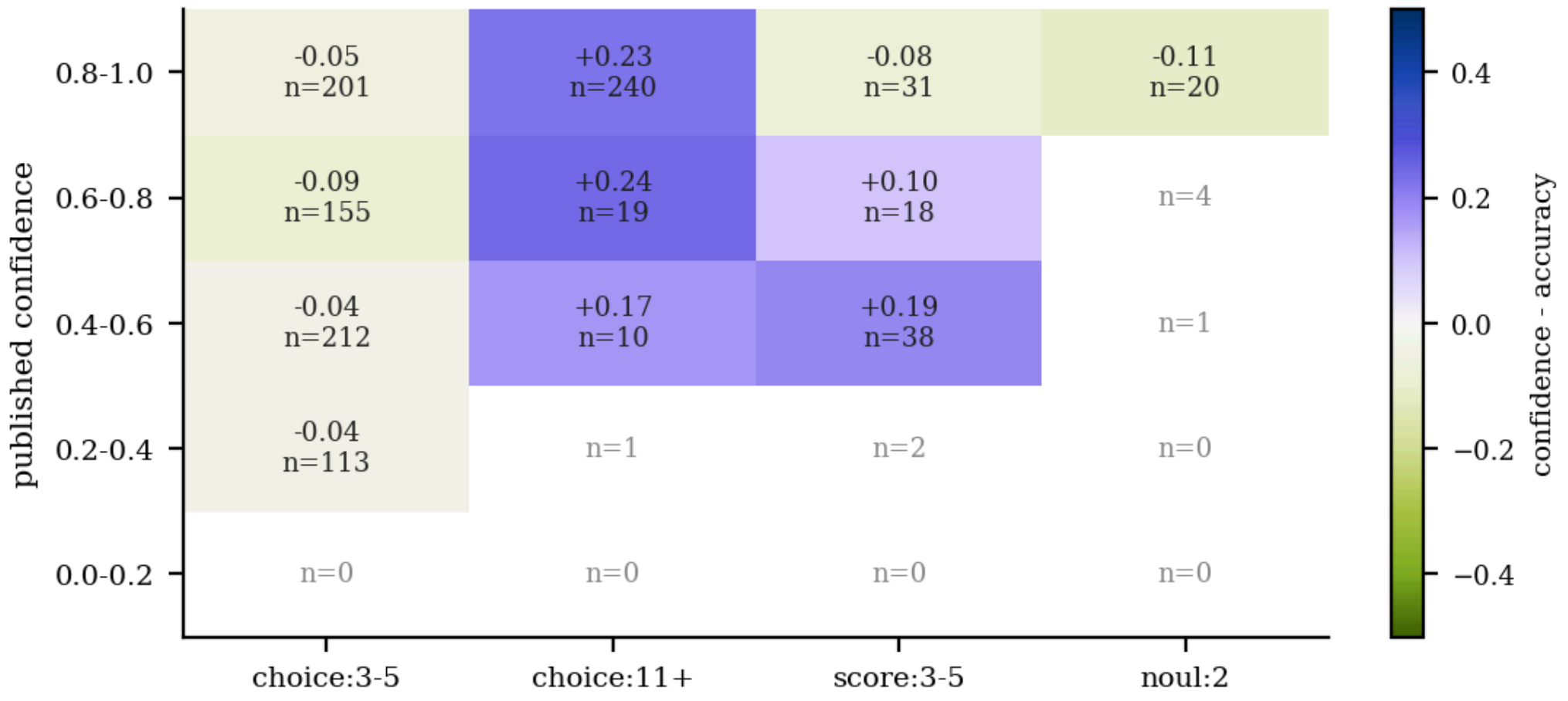

**Figure 7.** Where in the confidence range each bucket goes wrong. Cells show mean confidence minus accuracy (red: over-confident) with the number of items; bins with fewer than five items show only the count.

## 6.3 Look one level below the bucket

Buckets are a convenient proxy for what drives over- or under-confidence: they group items by *shape* (type and option count), while the cause may lie in the *content*. Figure 8 splits the pool by data category. Inside the single `choice:3-5` bucket the gap runs from -0.22 (ml_Japanese) to +0.15 (ml_Arabic), so one temperature per bucket cannot make every category right. It fixes the average only, so treat it as a first step.

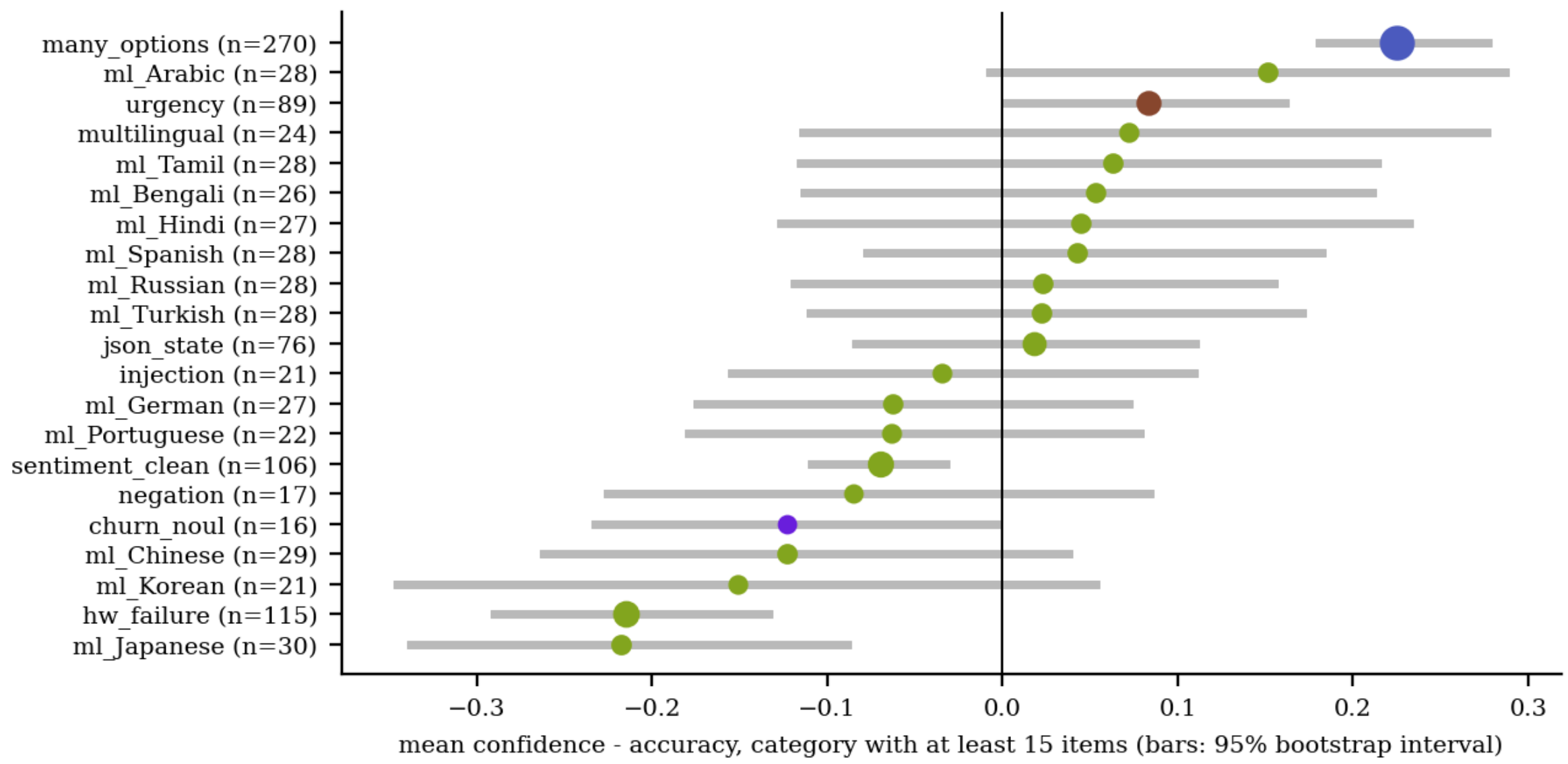

**Figure 8.** Slicing below the bucket: signed gap per data category. Marker size grows with n; colour is the category's main bucket (same colour code as Figure 4). Inside `choice:3-5` the categories differ in sign, which is a hint that a bucket is a coarse proxy for what really drives over- or under-confidence.

> **Check yourself.** Take the row for `noul:2` in Table 5: n = 25, accuracy 96%, mean confidence 0.84. Is it under-confident? The sign says yes (gap -0.117), but its ECE 0.154 is below the chance floor of 0.195, so with 25 items you should not trust the direction.

## 7 Step 3: fit one temperature per bucket, and the alternatives

### 7.1 Temperature scaling, bucket by bucket

Temperature scaling [Guo et al. 2017] has one parameter, so it needs few items and cannot reorder anything: dividing logits by a positive number never changes which option is largest, so accuracy is untouched and only the confidence changes. If you log the raw option logits, divide them by T and re-apply softmax. If you only logged the *published* probabilities, as we do here, raise them to the power 1/T and renormalise. This is the same operation, and the T you fit is then a multiplier on top of the temperature Laya already applied. Listing 2 evaluates every candidate T for every item once and stores the result in two tables, so that every later fit, cross-fit or bootstrap is just a sum over rows. Guo et al. fit one temperature per model; fitting one per bucket is the only change here.

```
GRID = np.geomspace(0.2, 20, 60)                      # candidate temperatures

def temper_tables(probs, gold):
    """For every item and every T in GRID: NLL of the gold option, and the top probability, of p**(1/T)."""
    nll, top = np.empty((len(probs), len(GRID))), np.empty((len(probs), len(GRID)))
    for i, (p, g) in enumerate(zip(probs, gold)):
        z = np.log(np.asarray(p) + 1e-9)[None, :] / GRID[:, None]
        z -= z.max(axis=1, keepdims=True)              # the largest logit is now 0
        lse = np.log(np.exp(z).sum(axis=1))
        nll[i], top[i] = lse - z[:, g], np.exp(-lse)
    return nll, top

def fit_temperature(nll_rows):
    """The T that minimises the summed NLL of the rows you pass in (one bucket, one fold, ...)."""
    return GRID[nll_rows.sum(axis=0).argmin()]
```

**Listing 2.** A grid search for the temperature. `nll` and `top` have one row per item and one column per candidate T. The grid is coarse (about 8% steps), which is finer than the sampling noise in the estimate.

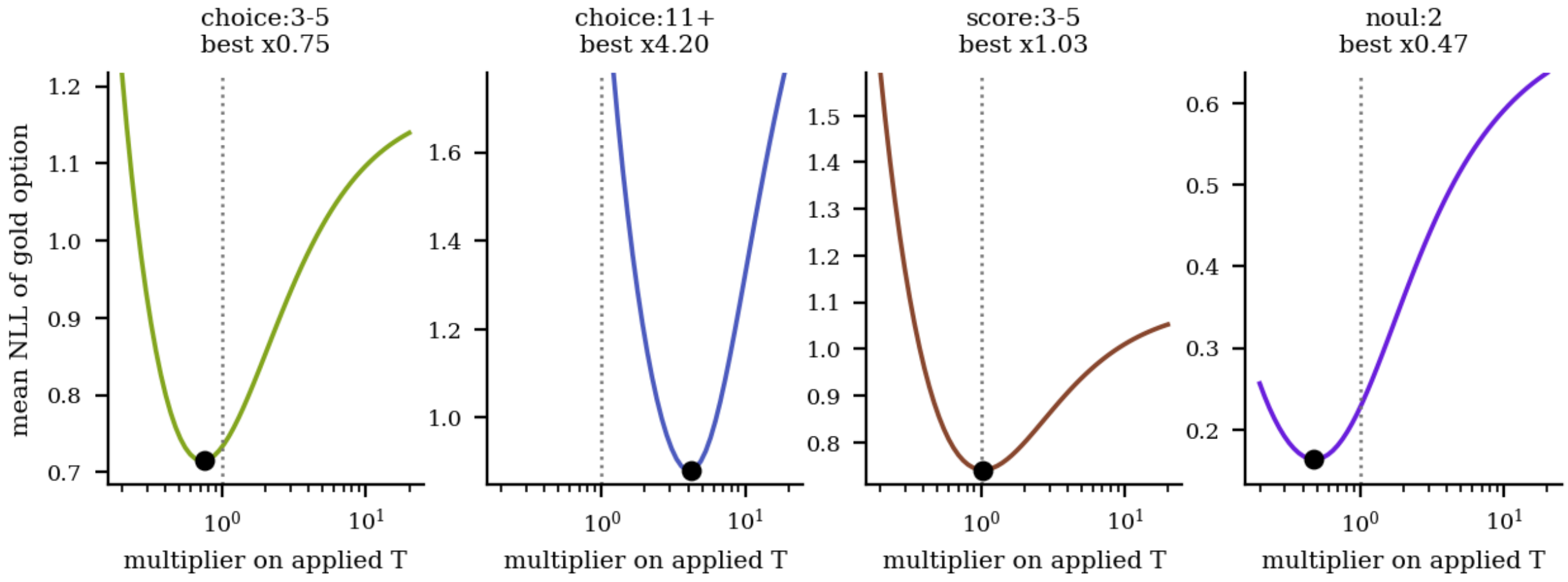

**Figure 9.** Step 3a: the temperature search. Each panel is one bucket; the curve is the mean negative log-likelihood of the correct option when the published probabilities are re-tempered by the multiplier on the x axis. The black dot is the minimum, the dotted line is 'leave the published temperature as it is'.

**Table 6.** Fitted temperature multipliers per bucket on all items (95% bootstrap interval, 300 resamples) and what they imply for the total temperature that should have been applied to the raw logits. A multiplier above 1 softens, below 1 sharpens.

| Bucket | n | Multiplier T | 95% interval | Applied T (stored, clamped) | Implied total T | ECE: published | ECE: cross-fitted |
|---|---|---|---|---|---|---|---|
| choice:3-5 | 681 | 0.75 | 0.64-0.82 | 1.760 | 1.33 | 0.054 | 0.051 |
| choice:11+ | 270 | 4.20 | 3.59-4.54 | 0.500 | 2.10 | 0.225 | 0.059 |
| score:3-5 | 89 | 1.03 | 0.75-1.41 | 1.251 | 1.29 | 0.137 | 0.116 |
| noul:2 | 25 | 0.47 | 0.20-0.88 | 1.983 | 0.94 | 0.154 | 0.068 |

Read Table 6 bucket by bucket. choice:11+ needs a multiplier of 4.20 (3.59-4.54): its published probabilities must be flattened considerably. Because the runtime already applied a temperature of 0.50 to it, the implied *total* temperature is 2.10, against the stored value of 0.1006. In the other direction, `choice:3-5` needs 0.75, a sharpening. The intervals for the small buckets are wide (0.20-0.88 for `noul:2`, which even touches the lower end of the grid), a first warning that the 25-item bucket should not be trusted to the second decimal.

The multiplier for choice:11+ also has a data-handling caveat. The logged probabilities are rounded to four decimals, and 99% of the choice:11+ items have at least one option at exactly 0, which `temper_tables` floors at 1e-9 before raising to a power. That floor is a choice. With a floor of 1e-6 the multiplier for choice:11+ is 2.84, and with 5e-5 (half the rounding step) it is 2.25, against 4.20. The cross-fitted macro-ECE hardly moves (0.076-0.077 over 10 fold assignments), so the repair holds, but the size of the flattening depends on it. A multiplier fitted on rounded probabilities of saturated items only approximates one fitted on raw logits, so log the logits when you can.

> **Check yourself.** The multiplier for `choice:3-5` is below 1 although the temperature was already above 1 in the config (1.76). Does that mean the stored value is wrong? Not necessarily: it says that on *these* items, which differ from the data the table was fitted on, the same sharpening would fit better. The stored value is right for its own data and wrong for yours; that is why you refit.

## 7.2 Maps that work on the confidence alone

Sometimes you only have the reported confidence and the outcome, not the probability vector (for example, when you inherit a log). Three standard tools take just those two columns, and Listing 3 implements each in numpy. Histogram binning [Zadrozny and Elkan 2002] replaces every confidence by the accuracy of the (equal-mass) bin it falls in. Isotonic regression [Niculescu-Mizil and Caruana 2005] finds the best non-decreasing step function by pooling adjacent violators. Platt scaling [Platt 1999] fits a sigmoid to the logit of the confidence, i.e. two parameters.

```
def fit_isotonic(conf, ok):
    """Pool-adjacent-violators: the best non-decreasing map from confidence to accuracy."""
    xs, inv = np.unique(np.asarray(conf, float), return_inverse=True)      # tied confidences start in one block
    hits, count = np.bincount(inv, weights=np.asarray(ok, float)), np.bincount(inv).astype(float)
    blocks = [[h, c, x] for h, c, x in zip(hits, count, xs)]               # [sum of ok, count, right edge]
    i = 0
    while i < len(blocks) - 1:
        if blocks[i][0] / blocks[i][1] > blocks[i + 1][0] / blocks[i + 1][1]:
```

```
            blocks[i] = [blocks[i][0] + blocks[i + 1][0], blocks[i][1] + blocks[i + 1][1], blocks[i + 1][2]]
            del blocks[i + 1]
            i = max(0, i - 1)
        else:
            i += 1
    return np.array([b[2] for b in blocks]), np.array([b[0] / b[1] for b in blocks])

def apply_isotonic(model, conf):
    edges, vals = model
    return vals[np.clip(np.searchsorted(edges, conf, side="left"), 0, len(vals) - 1)]

def fit_hist(conf, ok, bins=8):
    """Equal-mass histogram binning: every confidence becomes the accuracy of its bin."""
    bins = int(min(bins, max(2, len(conf) // 10)))
    edges = np.quantile(conf, np.linspace(0, 1, bins + 1))[1:-1]
    idx = np.digitize(conf, edges)
    return edges, np.array([ok[idx == b].mean() if (idx == b).any() else ok.mean() for b in range(bins)])

def apply_hist(model, conf):
    edges, vals = model
    return vals[np.digitize(conf, edges)]

def logit(c):
    c = np.clip(c, 1e-4, 1 - 1e-4)
    return np.log(c / (1 - c))

def fit_platt(conf, ok, iters=50, ridge=1e-3, max_step=1.0):
    """Platt scaling of the logit: accuracy ~ sigmoid(a * logit(conf) + b), Newton steps, ridge towards (1, 0)."""
    x, a, b = logit(conf), 1.0, 0.0
    for _ in range(iters):
        p = 1 / (1 + np.exp(-(a * x + b)))
        grad = np.array([np.sum((p - ok) * x), np.sum(p - ok)]) + ridge * np.array([a - 1, b])
        w = p * (1 - p) + 1e-6
        hess = np.array([[np.sum(w * x * x), np.sum(w * x)], [np.sum(w * x), np.sum(w)]]) + ridge * np.eye(2)
        step = np.clip(np.linalg.solve(hess, grad), -max_step, max_step)     # damped: a near-separable bucket would
otherwise diverge
        a, b = a - step[0], b - step[1]
        if np.abs(step).max() < 1e-6:
            break
    return a, b

def apply_platt(model, conf):
    a, b = model
    return 1 / (1 + np.exp(-(a * logit(conf) + b)))
```

**Listing 3.** Three confidence-only maps. Each `fit_*` returns parameters; each `apply_*` maps new confidences. The histogram guards its bin count by the sample size (at least ten items per bin); isotonic regression has no such guard, which matters for tiny buckets. The Newton iteration of `fit_platt` is damped (`max_step`): without it, a bucket whose confidences are nearly all 1 sends the slope off to huge values (see Section 12).

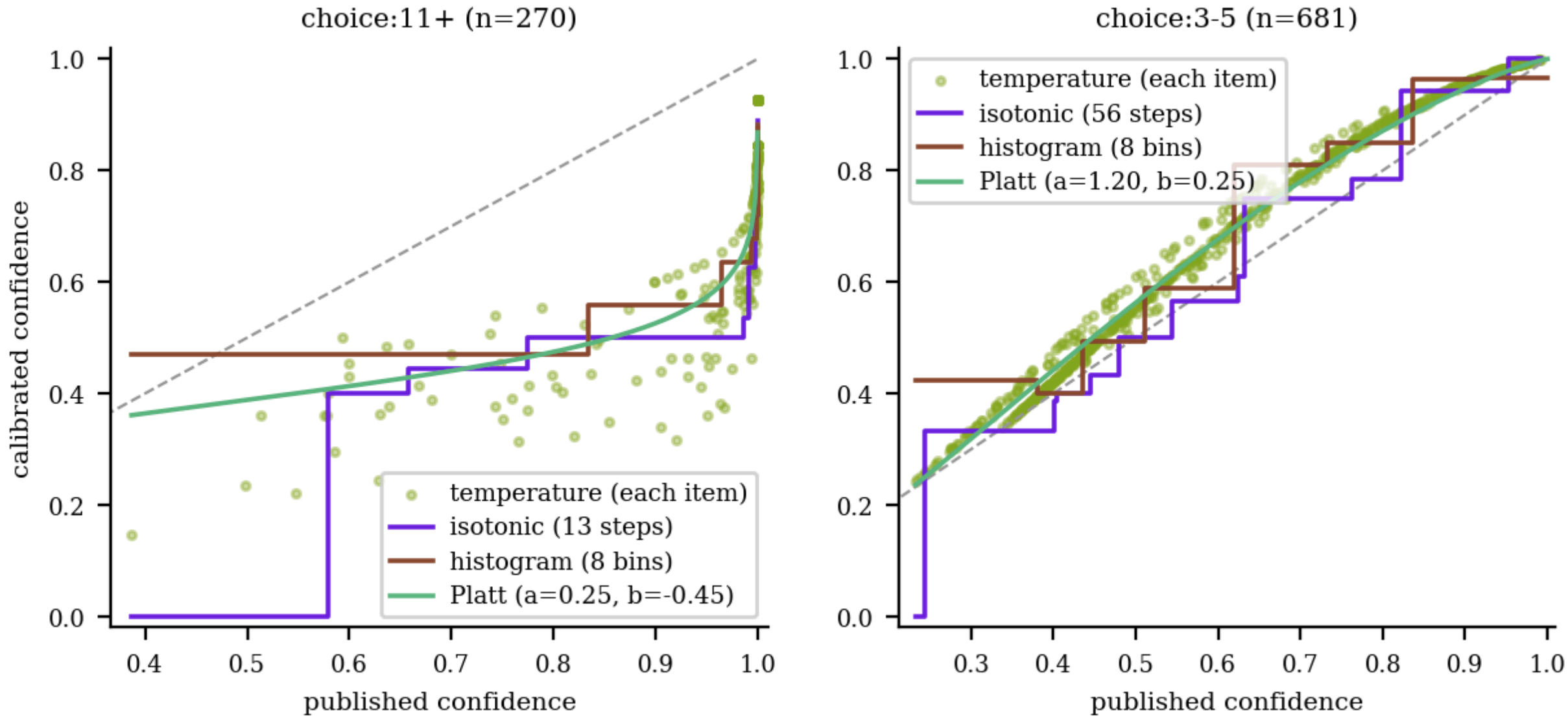

**Figure 10.** Step 3b: what each recalibration actually does to a confidence, fitted on every item of the bucket (an illustration; Step 4 fits them out-of-sample). Temperature moves items by an amount that depends on the whole probability vector, so it is a cloud rather than a curve; isotonic, histogram and Platt are one-to-one functions of the published confidence.

Figure 10 makes the differences concrete: in choice:11+ the median published confidence is 0.999, and temperature scaling moves the central 80% of its items to 0.41-0.93. Isotonic regression and the histogram, which can only reuse confidences that exist, send the top region to about 0.89, the accuracy of the most confident items. The Platt fit for that bucket has slope 0.25 and intercept -0.45: it compresses the logit scale and shifts it, the logit-space counterpart of the temperature's flattening. It is a smooth two-parameter curve, whereas the histogram and isotonic maps are steps that can only be as fine as the bucket's data allow.

## 8 Step 4: validate by cross-fitting, holding out a set, sizing the fit set and stressing the metric

A recalibration map is itself a fitted model, so it can overfit, and an ECE computed on the items it was fitted to is optimistic. The cure is cross-fitting: split the items into five folds, fit the map on four, apply it to the fifth, and rotate. Every item then receives a confidence from a map that never saw it, and you can still use all items for the final report. This is an in-pool check, because items from one generator sit on both sides of every split; Section 8.4 holds out a whole set. Listing 4 is the whole procedure, parametrised by a "fitter" so that every method in Step 3 plugs in.

```
def fold_ids(n, nfold=5, seed=0):
    return np.random.default_rng(seed).permutation(n) % nfold

# D is a dict of aligned arrays: conf (published top probability), ok (0/1), b (bucket name),
# nll and top (from temper_tables). A "fitter" takes (D, train_idx) and returns f(test_idx) -> calibrated confidence.
def by_temperature(D, tr):
    j = D["nll"][tr].sum(axis=0).argmin()
    return lambda te: D["top"][te, j]

def by_isotonic(D, tr):
    m = fit_isotonic(D["conf"][tr], D["ok"][tr])
    return lambda te: apply_isotonic(m, D["conf"][te])

def by_histogram(D, tr):
    m = fit_hist(D["conf"][tr], D["ok"][tr])
    return lambda te: apply_hist(m, D["conf"][te])

def by_platt(D, tr):
    m = fit_platt(D["conf"][tr], D["ok"][tr])
    return lambda te: apply_platt(m, D["conf"][te])

def crossfit(D, fitter, nfold=5, seed=0, min_n=10, per_bucket=True, max_fit=None):
    """Calibrated confidence for every item, each produced by a map fitted on the other folds.
    Buckets with fewer than min_n fitting items keep their published confidence."""
    n = len(D["ok"])
    fold, out = fold_ids(n, nfold, seed), D["conf"].copy()
    rng = np.random.default_rng(seed + 1000)
    for f in range(nfold):
        for u in (np.unique(D["b"]) if per_bucket else [None]):
            sel = (D["b"] == u) if per_bucket else np.ones(n, bool)
            tr, te = np.where(sel & (fold != f))[0], np.where(sel & (fold == f))[0]
            if max_fit and len(tr) > max_fit:
                tr = rng.choice(tr, max_fit, replace=False)
            if len(te) and len(tr) >= min_n:
                out[te] = fitter(D, tr)(te)
    return out
```

**Listing 4.** Cross-fitting per bucket. A bucket with fewer than `min_n` fitting items is left untouched, so it is not fitted to noise; `per_bucket=False` gives the one-global-temperature baseline; `max_fit` caps the fit-set size for the experiment in Figure 12.

### 8.1 Compare the methods

**Table 7.** All recalibration methods, five-fold cross-fitted on the 1,065 pooled items (every item is calibrated by a map fitted on the other four folds, bucket by bucket unless stated). Macro-ECE with a 95% bootstrap interval; "in-sample" fits and scores on the same items; the Brier score [Brier 1950] is for the top label; the last column repeats the cross-fit with 10 different fold assignments.

| Method | Pooled ECE | Macro-ECE [95% CI] | Macro-ECE in-sample | Macro-ECE, 10 fold seeds | Binary NLL | Brier (top label) |
|---|---|---|---|---|---|---|
| published confidence | 0.073 | 0.143 [0.127-0.184] | 0.143 | n/a | 0.694 | 0.184 |
| one global temperature | 0.173 | 0.210 [0.186-0.240] | n/a | 0.205 ± 0.002 | 0.660 | 0.219 |
| temperature per bucket | 0.039 | 0.073 [0.068-0.127] | 0.074 | 0.078 ± 0.006 | 0.508 | 0.169 |
| isotonic per bucket | 0.051 | 0.108 [0.076-0.175] | 0.000 | 0.074 ± 0.007 | 0.570 | 0.173 |
| histogram (8 bins) per bucket | 0.024 | 0.059 [0.057-0.112] | 0.000 | 0.076 ± 0.008 | 0.520 | 0.171 |
| Platt per bucket | 0.036 | 0.078 [0.061-0.155] | 0.035 | 0.046 ± 0.005 | 0.534 | 0.170 |

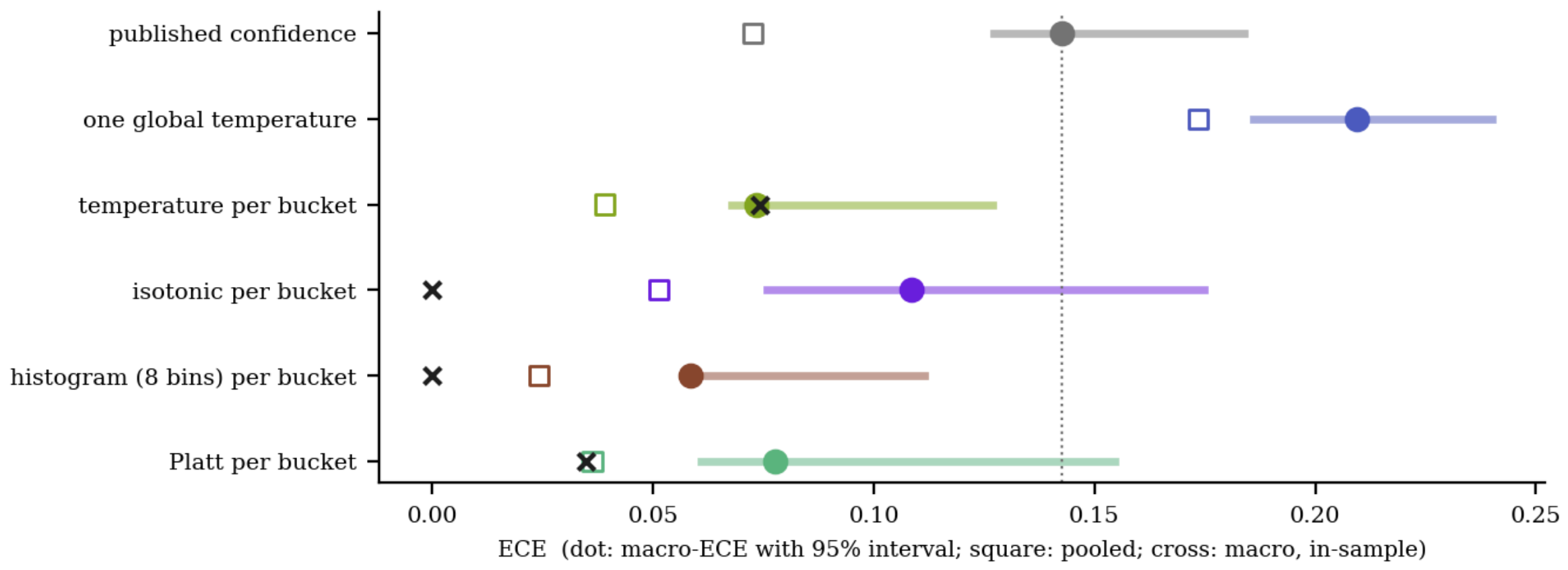

**Figure 11.** Step 4a: calibration error of each method, cross-fitted. A map that looks better in-sample (black cross) than out-of-sample (dot) is partly fitting noise; the distance between the two is that method's optimism.

The published confidences have macro-ECE 0.143, and one global temperature, fitted on all items, raises pooled ECE from 0.073 to 0.173 (macro-ECE 0.210): the single multiplier it uses (2.43) has to serve choice:11+, which wants 4.20, and `choice:3-5`, which wants 0.75. A temperature per bucket brings macro-ECE to 0.073 (0.068-0.127) and pooled ECE to 0.039; the paired reduction in macro-ECE is 0.069 (95% bootstrap interval 0.032-0.085). These intervals resample items after the cross-fit and so ignore the variance of the fitted maps; the fold-seed column is the better guide to stability.

Which per-bucket method is best depends on the fold assignment. On the first one the lowest macro-ECE belongs to the histogram map (0.059), but a single assignment is a noisy referee, partly because the 25-item `noul:2` bucket is re-cut every time. The last column of Table 7 repeats the cross-fit with 10 assignments: temperature 0.078, isotonic 0.074, histogram (8 bins) 0.076, Platt 0.046. On that measure the best is Platt, and the seed-to-seed standard deviation of the methods runs from 0.002 to 0.008. Optimism (out-of-sample minus in-sample macro-ECE) is -0.001 for the temperature, 0.059 for the histogram, 0.108 for isotonic regression and 0.043 for Platt scaling; the more parameters a map has, the more of its in-sample gain is noise. Platt scaling also has a trap: with a plain Newton iteration the slope of one nearly separable bucket runs away (up to |a| = 79820) and the cross-fitted macro-ECE is 0.139 instead of 0.078. Listing 3 damps the step for this reason.

## 8.2 How many labelled items do you need?

Figure 12 caps the number of fitting items per bucket, to see whether an afternoon of labelling is enough, and measures macro-ECE on the held-out folds. The temperature reaches 0.089 with only 10 items per bucket and 0.078 with 320, so it is almost fully fitted after a few dozen labels. The confidence-only maps need more: to get below 65% of the published macro-ECE, Platt scaling needs 20 items per bucket, the histogram 40, isotonic regression 80 and the temperature 10. Platt scaling overtakes the temperature from 40 items per bucket and ends at 0.046 with 320, against 0.078 for the temperature, which has plateaued. Averaged over 10 fold assignments with all the data, the difference is largest in choice:11+ (0.029 against 0.068) and in `score:3-5` (0.064 against 0.128); we did not test why. Figure 13 shows the other side of the same fact: the multiplier itself varies by 3% (relative standard deviation) across training folds in choice:11+ and by 29% in `noul:2`.

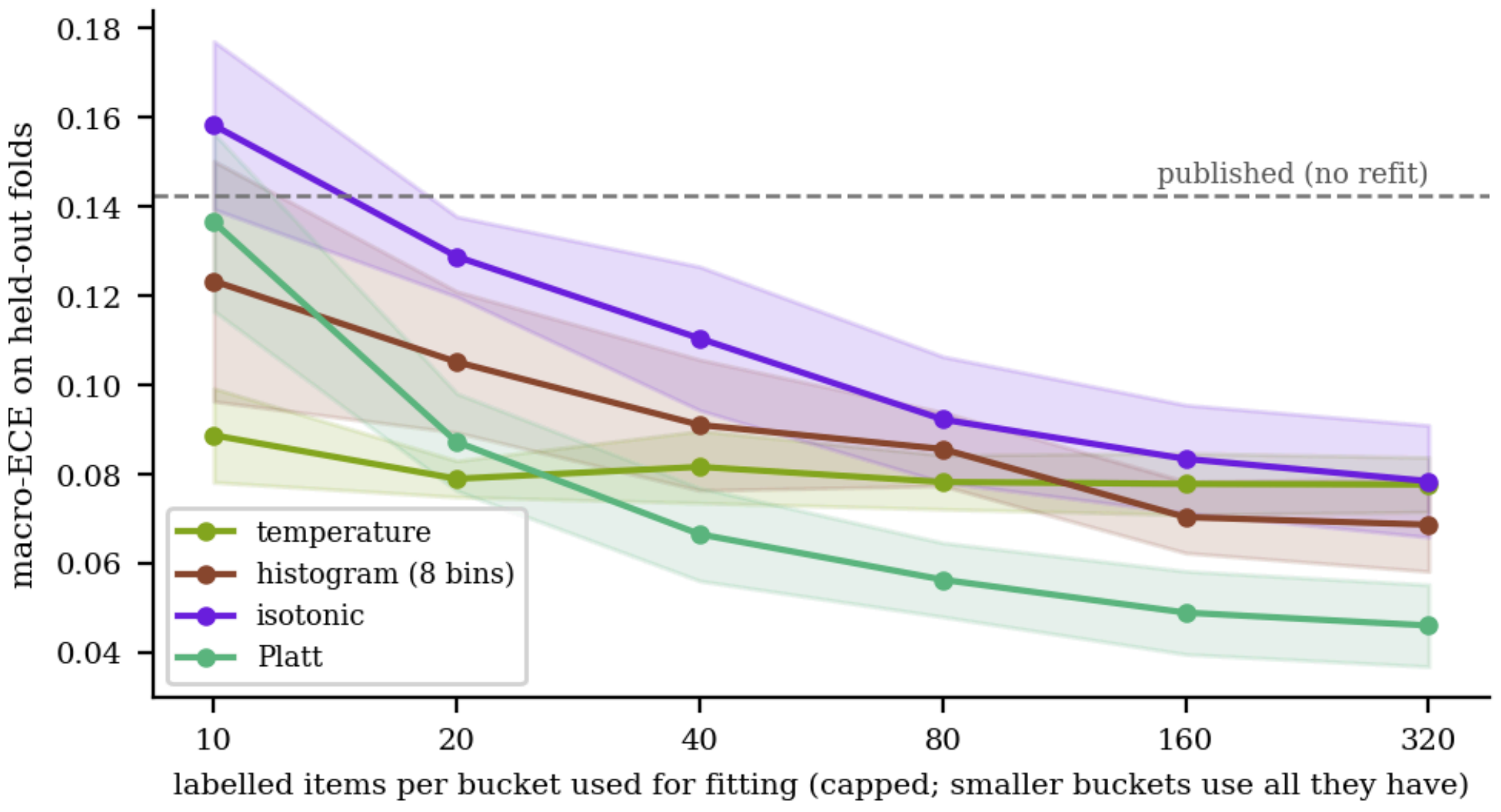

**Figure 12.** Step 4b: how many labelled items a method needs. Macro-ECE of the held-out folds as the number of fitting items per bucket is capped (mean and one standard deviation over 8 random draws). The one-parameter temperature is already close to its floor with the smallest fit sets; the confidence-only maps need more items before they catch up, and one of them overtakes it.

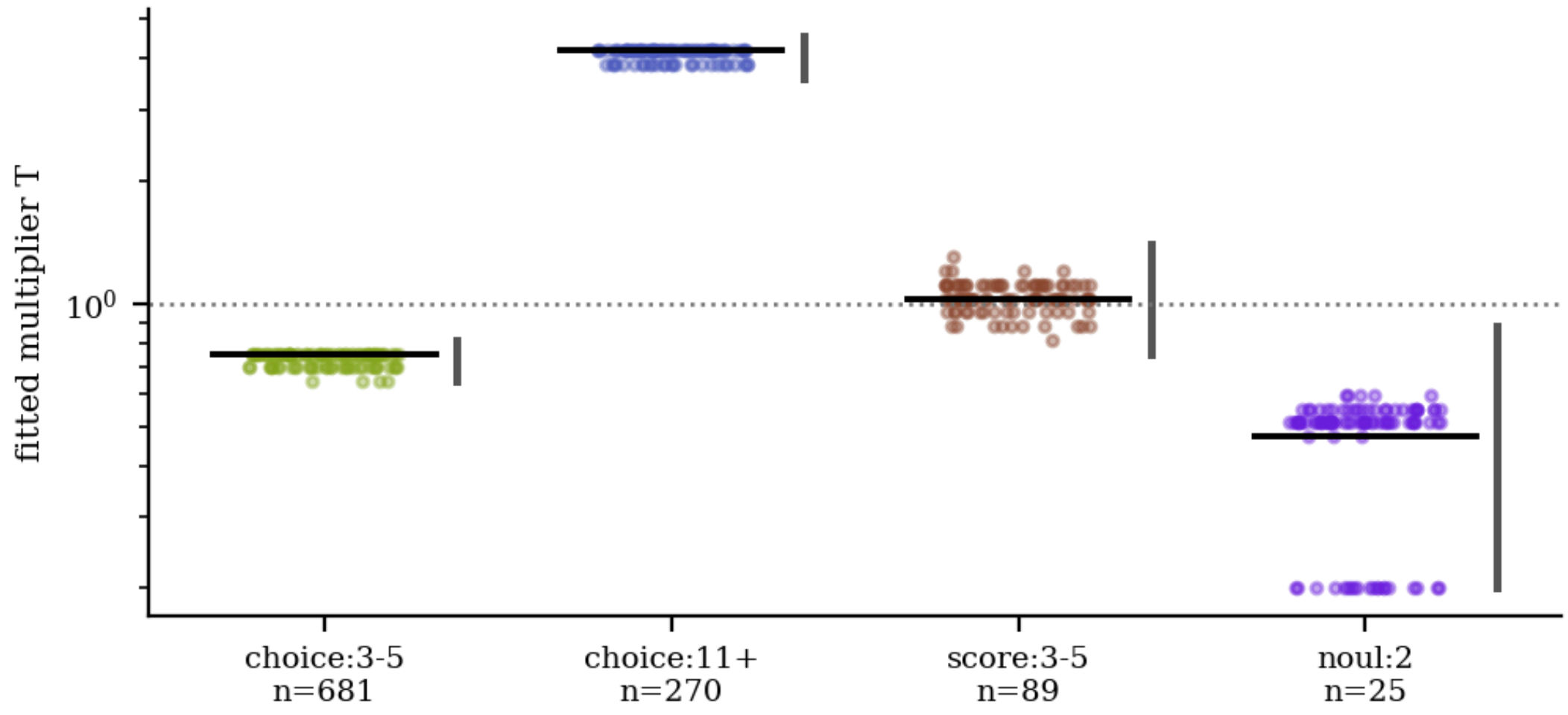

**Figure 13.** Step 4c: does the fitted temperature depend on which items you happened to use? Dots: the multiplier fitted on each of 100 training folds (20 fold assignments x 5); black tick: fit on all items; grey whisker: 95% bootstrap interval. Large buckets are stable; the 25-item bucket is not.

## 8.3 Stress the metric itself

Two checks test whether the result depends on a convention. First, ECE depends on the binning rule [Nixon et al. 2019]. Table 8 recomputes the per-bucket ECE with 5, 10, 15 and 20 equal-width bins and with 10 equal-mass bins. The absolute numbers move (the published confidences range from 0.133 to 0.156), but the ordering published > isotonic > temperature is the same under every rule. Second, the labels are LLM-made: Restricting to the 1028 of 1065 items on which an independent model agrees with the stored label gives macro-ECE 0.138 for the published confidences and 0.070 after the refit (all items: 0.143 and 0.073). The conclusion does not hinge on the items two LLMs disagree about.

**Table 8.** Does the bucket-macro ECE depend on the binning rule? Mean per-bucket ECE with five rule variants, for the published confidences and for two cross-fitted recalibrations.

| Binning rule | Published | Temperature per bucket | Isotonic per bucket |
|---|---|---|---|
| 5 equal-width bins | 0.133 | 0.065 | 0.096 |
| 10 equal-width bins | 0.143 | 0.073 | 0.108 |
| 15 equal-width bins | 0.149 | 0.085 | 0.111 |
| 20 equal-width bins | 0.156 | 0.090 | 0.116 |
| 10 equal-mass bins | 0.141 | 0.091 | 0.105 |

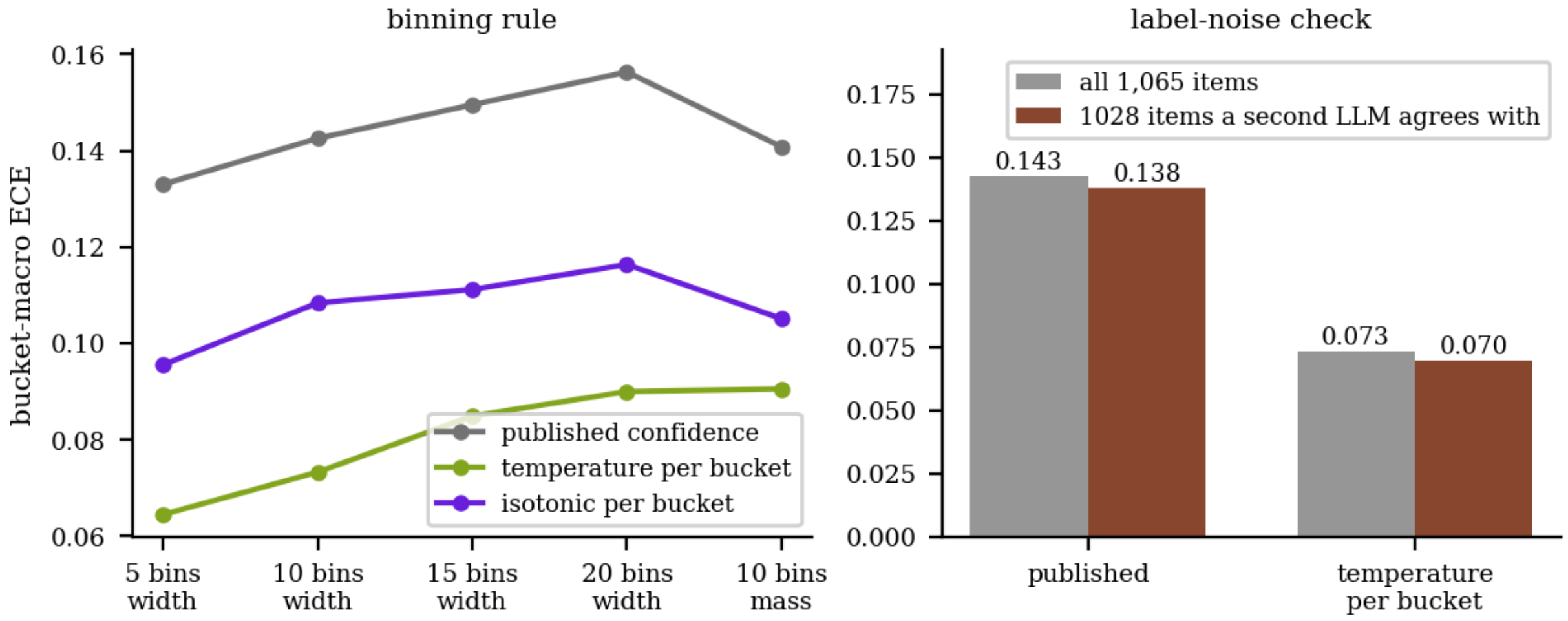

**Figure 14.** Step 4d: two checks that the conclusion does not depend on a convention. Left: bucket-macro ECE under five binning rules. Right: the same comparison on all items and on the subset where an independent LLM (gemma4-31b) agrees with the stored label.

### 8.4 Hold out a whole set

Random folds put items from the same generator, and some near-duplicates, on both sides of the split, so they favour the fit. A harder check fits the per-bucket temperature on six of the seven sets and applies it to the seventh, rotating through all seven. On this split the macro-ECE of the base checkpoint goes from 0.143 to 0.099 (random folds: 0.073), and pooled ECE from 0.073 to 0.037. The ECE falls on 5 of the 7 sets and rises on `ml_test` (0.058 to 0.092) and `urg_test` (0.143 to 0.156). So the per-bucket temperature carries over between sets of one generator only in part. Across the 33 checkpoints of Section 13 the set-held-out refit lowers macro-ECE for 21, against 31 with random folds. Read every cross-fitted number in this paper as an in-pool figure; the set-held-out one is closer to what new traffic will give.

> **Check yourself.** If two methods differ by less than their seed-to-seed standard deviation, which should you ship? The simpler one: here the per-bucket temperature, which also works with 10 items per bucket.

## 9 Step 5: deploy thresholds, abstention and cost

A calibrated confidence is only useful once something acts on it. Three decisions follow, and each can be checked the same way as the calibration itself: fit on some items, judge on others.

### 9.1 Write the temperatures back

The fitted multipliers are most useful if the model emits the corrected confidence itself, so that nothing downstream has to know about buckets. Listing 5 multiplies each multiplier into the stored temperature of its bucket (respecting the runtime's [0.5, 5] clamp) and returns a new config dictionary; the result for the base checkpoint is choice:3-5: 1.33, choice:11+: 2.10, score:3-5: 1.29, noul:2: 0.94. The same listing contains the gate used below: it abstains on a bucket that has no threshold instead of accepting everything, which is the safer failure mode.

```
QTYPES = ["choice", "score", "noul"]

def recalibrated_config(cfg, multiplier, lo=0.5, hi=5.0):
    """Fold per-bucket multipliers into the config's temperature table; the runtime clamps to [lo, hi] anyway."""
    table = dict(cfg.get("temperature_by_options", {}))
    for bucket, m in multiplier.items():
        stored = table.get(bucket, cfg["temperature"][QTYPES.index(bucket.split(":")[0])])
        table[bucket] = float(np.clip(np.clip(stored, lo, hi) * m, lo, hi))
    return {**cfg, "temperature_by_options": table}

def accept(answer_confidence, bucket, thresholds):
    """Gate one answer; a bucket without a threshold abstains rather than passing everything."""
    return answer_confidence >= thresholds.get(bucket, np.inf)
```

**Listing 5.** Writing the refit back into a config, and gating one answer. `accept(0.98, 'choice:11+', {'choice:11+': 0.97})` is True; for a bucket with no entry it is False.

## 9.2 Pick the cut-off on calibrated confidences, per what you can afford

A selective classifier answers when its confidence passes a bar [Chow 1970; Geifman and El-Yaniv 2017], and the relevant picture is the risk-coverage curve. TypedBench [Sharma et al. 2026] and Rafe and Das [2026] already evaluate selective prediction and gates for typed-decision models; what follows applies the same checks to Laya, per bucket. Figure 15 sorts all 1,065 items by confidence. Refitting leaves each bucket's own ordering unchanged (temperature is monotone) but changes how buckets interleave, which moves the area under the curve from 0.182 to 0.137. The usual rule "answer at 0.9 or above" answers 34.0% of items at 16.0% error on the published confidences, and 29.0% at 5.8% error after the refit: the cut-off means very different things before and after.

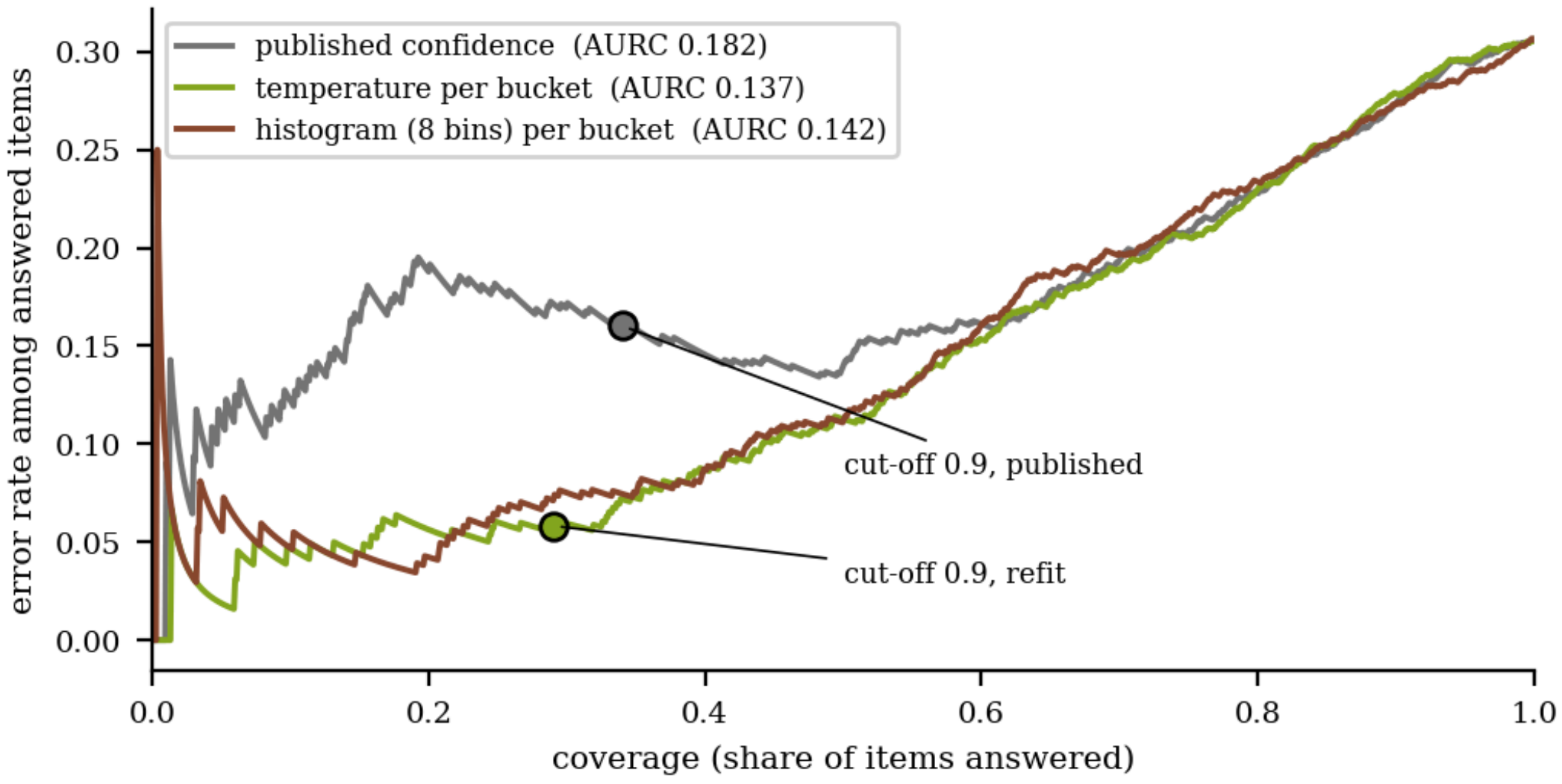

**Figure 15.** Step 5a: risk-coverage curves. Sorting items by confidence and answering the most confident first, the curve shows the error rate at each coverage. The dots mark what a fixed cut-off of 0.9 gives on the published and on the refit confidences.

A more useful way to set the bar is to pick the lowest cut-off that meets a target error, as in Listing 6, and to test it on held-out items; Table 9 reports the result for three targets. At a 10% target, a single cut-off on the published confidences answers only 6.9% of items and still misses the target (overall error 13.0%; target met in 0 of 10 fold assignments). The cut-off has to be high enough to exclude the confident-but-wrong choice:11+ items, so it excludes nearly everything else. The same rule on the refit confidences answers 46.3% at 10.5% error, and one cut-off per bucket on the published confidences answers 45.4% at 10.3%. Those two are close, so the gain comes from treating the buckets separately, and it does not need the temperature itself. Neither reliably meets the target: the overall error stays at or below 10% in 0 of 10 fold assignments for the refit cut-off and in 4 for per-bucket cut-offs.

```
def pick_threshold(conf, ok, target_error=0.10, min_accept=10):
    """Lowest confidence cut-off at which the accepted items are wrong at most target_error of the time."""
    conf, ok = np.asarray(conf, float), np.asarray(ok, float)
    o = np.argsort(-conf)
    err = 1 - np.cumsum(ok[o]) / np.arange(1, len(o) + 1)
    good = np.where((err <= target_error) & (np.arange(1, len(o) + 1) >= min_accept))[0]
    return conf[o][good.max()] if len(good) else np.inf    # inf: accept nothing

def best_cost_threshold(conf, ok, abstain_cost):
    """Cut-off minimising the mean cost when a wrong accepted answer costs 1 and an abstention costs abstain_cost."""
    conf, ok = np.asarray(conf, float), np.asarray(ok, float)
    cuts = np.unique(np.concatenate([conf, [np.inf]]))
    costs = [np.mean(np.where(conf >= t, 1 - ok, abstain_cost)) for t in cuts]
    return cuts[int(np.argmin(costs))]
```

**Listing 6.** Choosing a cut-off for a target error rate, and for a cost model in which an abstention costs a fixed fraction of a wrong answer.

**Table 9.** Gates aimed at a target error rate of 5%, 10% and 20%. Cut-offs are fitted on four folds and applied to the fifth; averages over 10 fold assignments. "Kept" counts the fold assignments in which the overall error of the accepted items stayed at or below the target; the last columns give the error among accepted items per bucket.

| Target | Gate | Coverage | Error | Kept | choice:3-5 | choice:11+ | score:3-5 | noul:2 |
|---|---|---|---|---|---|---|---|---|
| 5% | one cut-off, published confidence | 3.0% | 11.8% | 0/10 | n/a | 0.12 | n/a | 0.00 |
| | one cut-off, refit confidence | 26.0% | 6.0% | 0/10 | 0.04 | 0.12 | 0.01 | 0.05 |
| | a cut-off per bucket, published | 31.0% | 6.1% | 0/10 | 0.05 | 0.13 | 0.06 | 0.05 |

| Target | Gate | Coverage | Error | Kept | choice:3-5 | choice:11+ | score:3-5 | noul:2 |
|---|---|---|---|---|---|---|---|---|
| 10% | one cut-off, published confidence | 6.9% | 13.0% | 0/10 | n/a | 0.13 | n/a | 0.00 |
| | one cut-off, refit confidence | 46.3% | 10.5% | 0/10 | 0.09 | 0.15 | 0.07 | 0.04 |
| | a cut-off per bucket, published | 45.4% | 10.3% | 4/10 | 0.10 | 0.12 | 0.12 | 0.04 |
| 20% | one cut-off, published confidence | 72.1% | 20.1% | 4/10 | 0.16 | 0.27 | 0.23 | 0.04 |
| | one cut-off, refit confidence | 73.5% | 20.3% | 3/10 | 0.19 | 0.23 | 0.23 | 0.04 |
| | a cut-off per bucket, published | 70.9% | 19.9% | 6/10 | 0.20 | 0.20 | 0.21 | 0.04 |

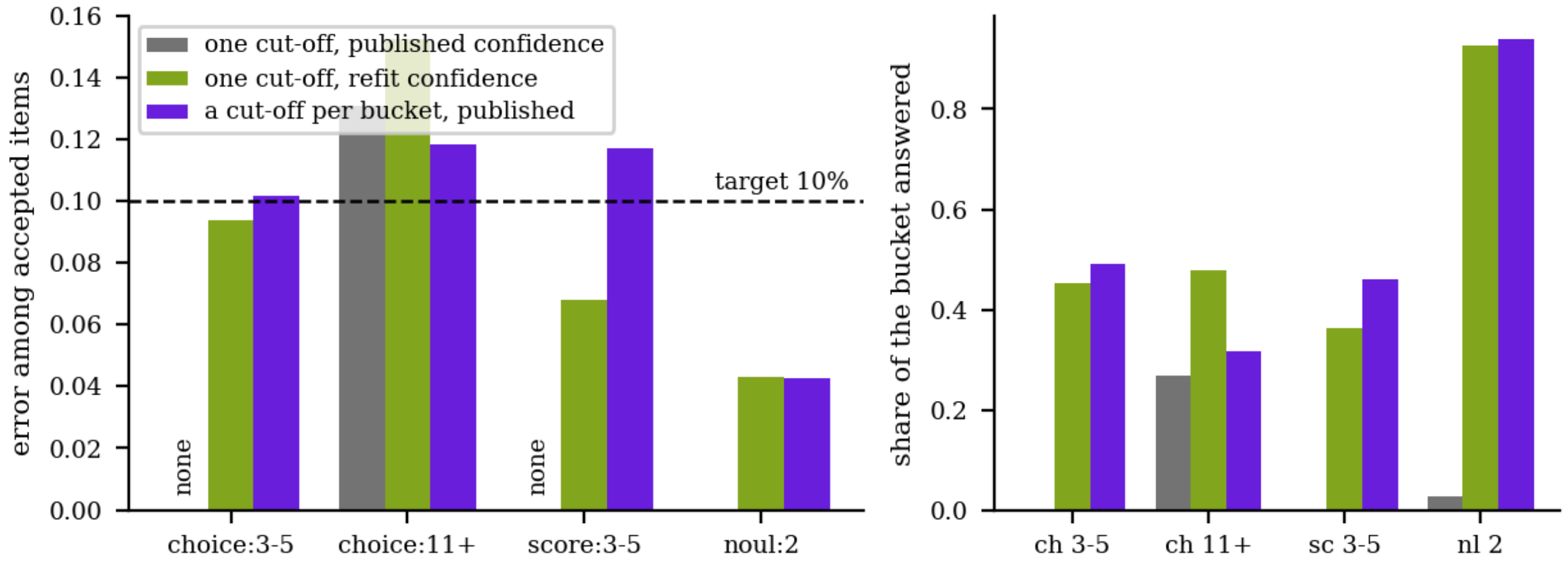

**Figure 16.** Step 5b: does the gate keep its promise in every bucket? Cut-offs aimed at a 10% error rate are chosen on four folds and evaluated on the fifth (mean of 10 fold assignments). Left: error among accepted items per bucket (‘none’: nothing accepted); right: share of each bucket that is answered.

Figure 16 shows where the promise breaks: with the single cut-off on published confidences, the accepted items in choice:11+ have an error of 0.13 against a target of 0.10, and the other buckets answer 0.1% of their items between them. Per-bucket cut-offs, or one cut-off on refit confidences, answer far more of the other buckets. Neither is perfect: choice:11+ still ends at 0.15 with the refit single cut-off and 0.12 with per-bucket cut-offs. A temperature cannot separate a confidently wrong item from a confidently right one in the same bucket; only a better ordering inside the bucket could.

## 9.3 Put a price on abstaining

If you can say what a wrong answer costs relative to a human review, you can skip the target error and minimise cost directly. Figure 17 assumes a wrong accepted answer costs 1 and an abstention costs c, so answering everything costs 0.306 per item. The refit confidences with one cut-off save up to 0.028 per item over the naive rule (largest at c = 0.15), and per-bucket cut-offs on the published confidences save up to 0.028. Both recover most of the same gain (they are within 0.001 of each other at best). The choice between “calibrate, then one cut-off” and “leave it alone, one cut-off per bucket” is therefore operational: the first keeps a single number for the whole system and the second needs a table.

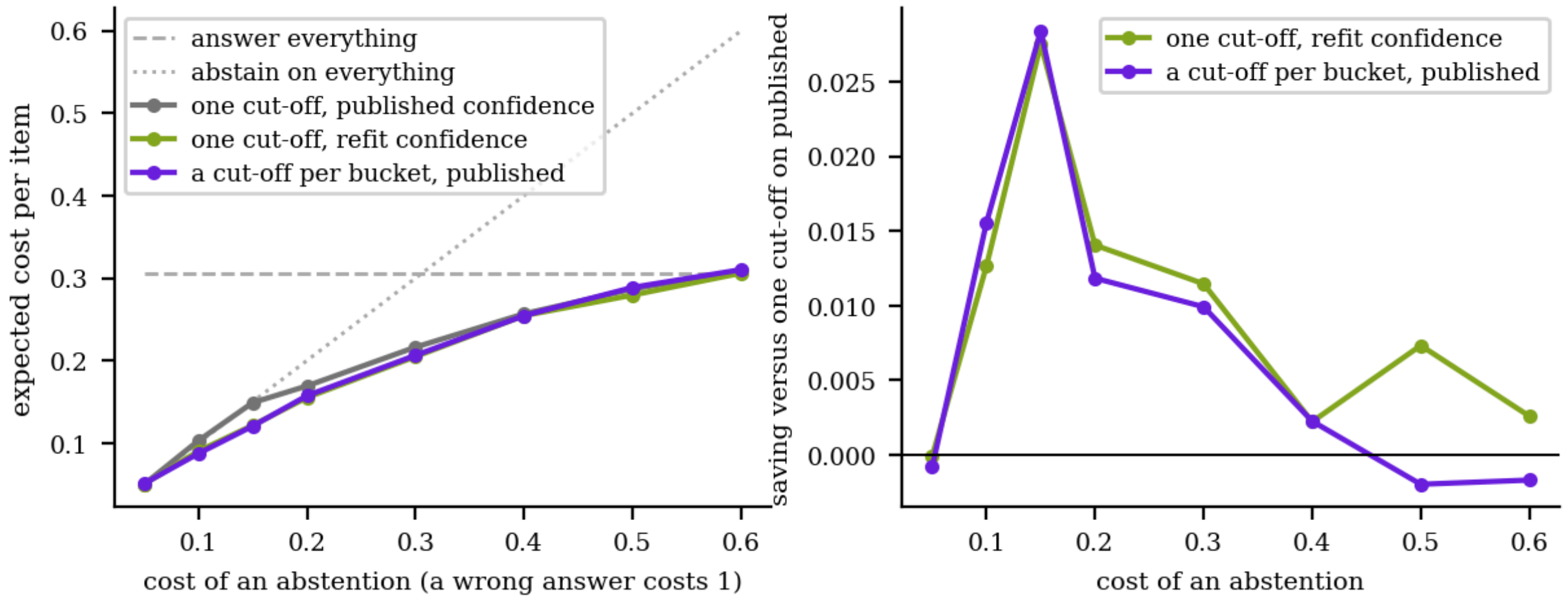

**Figure 17.** Step 5c: expected cost when answers below a cut-off are handed to a person. The cut-off is chosen on four folds to minimise cost and judged on the fifth. Right: how much each strategy saves over the naive single cut-off on published confidences.

**Check yourself.** Why does a per-bucket cut-off on the *published* confidence perform like one global cut-off on the *refit* confidence? Because a monotone map within each bucket does not change the order of items inside it; refitting only re-labels the scale, so any single cut-off on the new scale corresponds to a different cut-off in each bucket on the old one.

## 10 The five steps on one page, and one item followed through

Table 10 collects the answers in one place; it is the table you would paste into a ticket after doing this on your own model.

**Table 10.** The whole recipe on the worked example, one row per step: what was asked, and the number that answered it.

| Step | Question | Answer on the base checkpoint |
|---|---|---|
| 1 Measure | Is the pooled confidence calibrated? | Pooled ECE 0.073; bucket-macro ECE 0.143 (0.127-0.184). |
| 2 Diagnose | Who is wrong, in which direction? | choice:3-5 -0.05, choice:11+ +0.22, score:3-5 +0.08, noul:2 -0.12 (confidence minus accuracy). |
| 3 Fit | How much to flatten or sharpen? | choice:3-5 ×0.75, choice:11+ ×4.20, score:3-5 ×1.03, noul:2 ×0.47. |
| 4 Validate | Does it hold out of sample? | Macro-ECE 0.143 → 0.073 cross-fitted; 10 fold seeds 0.078 ± 0.006; stable with 10 items per bucket; clean-label subset 0.138 → 0.070. |
| 5 Deploy | What does the gate deliver? | 10%-error target: coverage 6.9% (error 13.0%) → 46.3% (error 10.5%); one cut-off per bucket on the published confidences gives 45.4% (error 10.3%); cost at c=0.1: 0.103 → 0.090. |

To make the effect concrete, follow individual items. Table 11 lists five mistakes at different published confidence levels. The refit moves each one according to its bucket: in a bucket that needs flattening the confidence goes down, in a bucket that needs sharpening it goes up, so the 0.9 gate changes its decision for 2 of the five. At the very top of choice:11+ the picture is different: 184 items are reported at 0.99 or above and only 82.1% of them are correct, while the cross-fitted confidence for the same items averages 0.830, still above the observed accuracy of the group. A temperature has one degree of freedom and cannot place the whole top of the distribution exactly; a confidence-only map on top of it (Section 7) can reduce that residual if the bucket has enough items.

**Table 11.** Five mistakes of the base checkpoint at different published confidence levels (from about 1.0 down to about 0.55), followed through the recipe. 'Refit' uses the multiplier fitted on all items of the item's bucket (Table 6).

| Set / bucket | Item (start of text) | Predicted | Gold | Published conf. | p(gold) | Refit conf. | Refit p(gold) | Accepted at 0.9: before / after |
|---|---|---|---|---|---|---|---|---|
| many_test / choice:11+ | Rate limiting errors (429) are hitting us randomly even though we're w... | latency | api | 0.999 | 0.0010 | 0.791 | 0.1526 | yes / no |
| probe / choice:11+ | The 'disconnected' error appears sporadically in our production enviro... | security | api | 0.952 | 0.0100 | 0.535 | 0.1809 | yes / no |
| many_test / choice:11+ | The product arrived broken. I have already picked up the return courie... | returns | refund | 0.842 | 0.1510 | 0.488 | 0.3243 | no / no |
| json_test / choice:3-5 | {"id": "TKT-442", "customer": "Bob Smith", "subject": "Request for API... | technical | other | 0.702 | 0.2278 | 0.787 | 0.1767 | no / no |
| urg_test / score:3-5 | I found a typo on the 'About Us' page. It says 'Company History' inste... | soon | not urgent | 0.552 | 0.2575 | 0.545 | 0.2602 | no / no |

**Check yourself.** Find the rows of Table 11 whose decision at 0.9 changes after the refit. Are the changes in the direction you would expect from the multipliers of Table 6? A bucket with a multiplier above 1 loses confidence, one below 1 gains it. Calibration improves the averages; it does not remove confidently wrong answers.

## 11 Choosing a method: a decision tree

The choice of method depends on what you logged and how many labelled items each bucket has. Figure 18 turns the experiments of Step 4 into a flow chart.

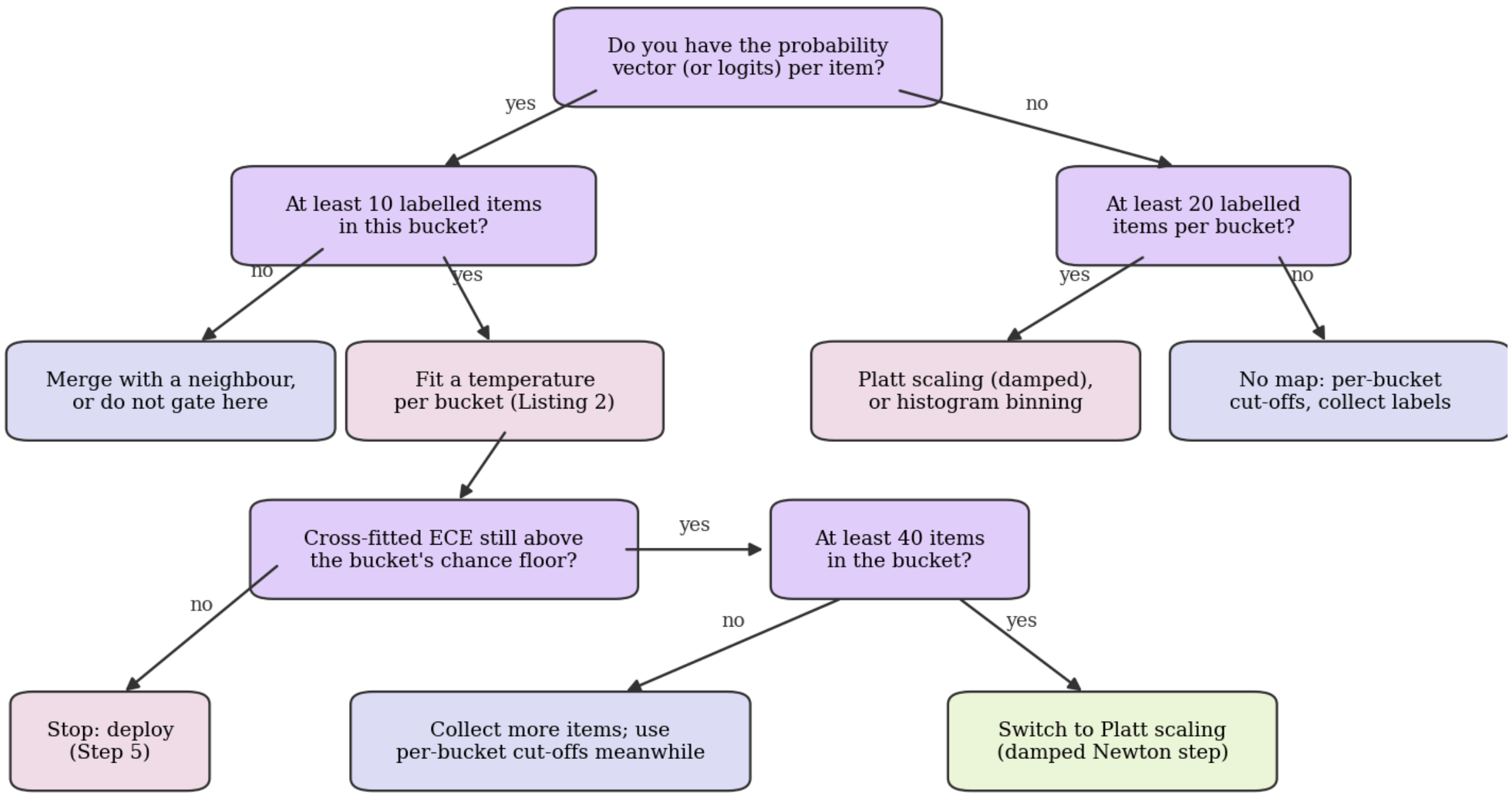

**Figure 18.** Choosing a method. Item counts in the boxes come from Figure 12: the size from which Platt scaling beats the temperature (left), and the size at which Platt scaling first reaches 65% of the published macro-ECE (right). Treat them as order-of-magnitude guides for this data, not constants.

Start with the temperature: it is the only method that is already close to its floor with 10 items per bucket, it never changes the ordering of the options, and it can be written back into the model's own config, so downstream code needs no knowledge of the recalibration. Move to a confidence-only map when the probabilities are not available (a log of replies, say), or when the residual after the temperature is still above the chance floor of Step 2 *and* the bucket has enough items. In this example that means Platt scaling from about 40 items per bucket (0.046 against 0.078 for the temperature over 10 fold assignments); histogram binning and isotonic regression need more items still, and isotonic regression needs clipping on small buckets.

## 12 Pitfalls: clamps, retrained checkpoints, small buckets, label noise and shift

### 12.1 The stored table is not what runs

The base config stores 1 entry that the runtime does not apply as written: the `choice:11+` value 0.1006 is clamped to 0.50. The package's source explains why: a temperature that small multiplies the logits roughly tenfold and would report a coin flip as a near-certainty, so the runtime refuses it. The clamp only *bounds* the damage, though, and a factor-two sharpening is still applied to the bucket that needs flattening, by a factor between about 2.2 and 4.2 depending on how the zeros in the logged probabilities are floored (Section 7). If you fit your multiplier on the *published* probabilities, as in this tutorial, the clamp is already accounted for. If you fit on raw logits, remember that the value you write back is clamped too.

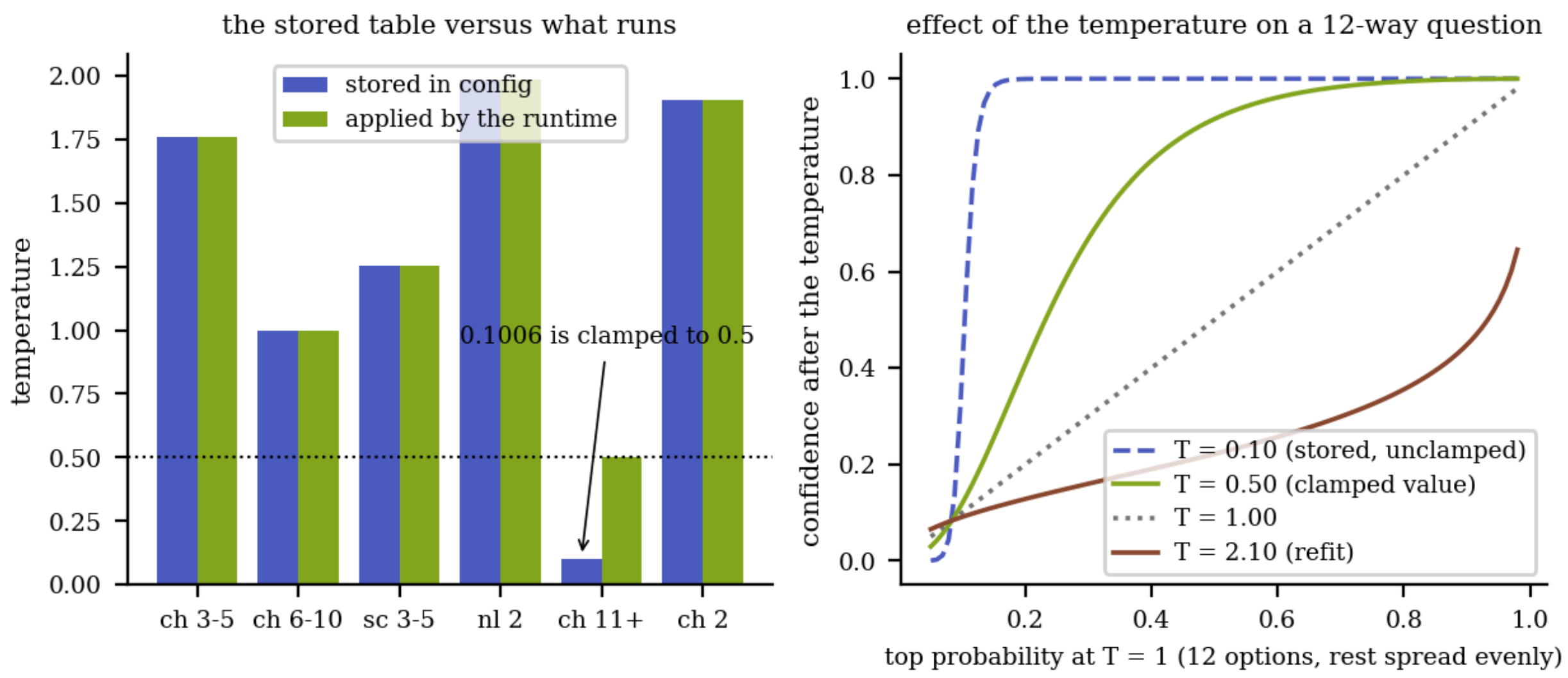

**Figure 19.** Pitfall 1: the stored table is not what runs. Left: stored and applied temperature per bucket; the 11+ entry of 0.1006 would sharpen by a factor of ten, so the runtime clamps it to 0.5. Right: what a temperature does to the top probability of a 12-option question (the remaining mass spread evenly), including the refit value for this bucket.

### 12.2 The retrained checkpoints here have no per-bucket table

This subsection reports what the saved configurations of the checkpoints in this study contain; because we did not vary the trainer's settings, it says nothing about Laya's trainer in general. All 30 checkpoints we retrained carry three per-type temperatures and no per-bucket table (0 of 30 have one). For 25 of 30 the choice temperature sits exactly on the ceiling of 5.0, and 27 of 30 have at least one of the three values on a clamp boundary. The trainer's own report agrees for the main retrained model: it reports "the fitted temperature 5 is on the [0.5, 5] clamp", and for the score and noul types it warns that the fit used only 16 and 11 calibration items. The package's fitting helpers have sample floors (Table 12), and the per-bucket temperature helper needs 2000 items per bucket, more than any bucket of this study has. That would explain the missing table, but we did not rerun the trainer with larger calibration sets to confirm it. A numpy refit with the floor of ten items used in Listing 4 avoids the floor.

**Table 12.** The sample floors built into the Laya package's own calibration helpers, against the bucket sizes of the worked example. A bucket below a floor is silently omitted by the helper (no entry is written).

| Bucket | Items | temperature per bucket (fit_temperature_map): floor 2000 | histogram map (fit_binning_map): floor 200 | per-bucket cut-off (fit_abstention_thresholds): floor 100 |
|---|---|---|---|---|
| choice:3-5 | 681 | below | meets | meets |
| choice:11+ | 270 | below | meets | meets |
| score:3-5 | 89 | below | below | below |
| noul:2 | 25 | below | below | below |

Across the 30 retrained checkpoints with enough items, the stored value is also a poor guide to what a refit asks for: whatever temperature the trainer stored, the multiplier for `choice:3-5` runs from 1.03 to 2.08. Refit each checkpoint you intend to gate on; do not copy a table from another.

### 12.3 Small buckets and confidence-only maps

In `noul:2` (25 items) the temperature interval is 0.20-0.88 and the chance floor of the ECE is 0.20. The cross-fitted isotonic map emits exactly 0 or 1 for 100% of the bucket's items (7% in choice:3-5, 0% in choice:11+, 40% in score:3-5). An isotonic step function with a block of all-correct or all-wrong items has no better output than a probability of one or zero, and a downstream gate treats that as certainty. This is why the macro-ECE of isotonic regression in Table 7 is worse than a quick look at the large buckets suggests. Clip the output, or choose the temperature for tiny buckets, and prefer to withhold gating on a bucket you cannot calibrate.

### 12.4 Label noise

Labels written by an LLM carry their own errors. Section 8 showed that the conclusion survives when the 37 items on which a second model disagrees are removed. In your own data, the same check is cheap: re-label a sample independently, drop the disagreements, and see whether the macro-ECE and the multipliers move. If they move a lot, calibrate against the cleaner subset: labels that are wrong lower the measured accuracy, so a fit on them pulls the confidence below what the clean labels would justify.

### 12.5 Shift: a map is a property of the data it was fitted on

Calibration measured on one text distribution may not hold on another, as Ovadia et al. [2019] and Santillana [2026] also report for other models. We perturb four sets under six conditions (typos at two rates, upper case, lower case without punctuation, truncation to the first half, tone stripped) and ask how much a histogram map fitted on the clean text helps. The accuracy of the base checkpoint drops from 68% on clean text to 50% under `upper`, and its macro-ECE rises from 0.170 to 0.260. The clean-text map brings the latter to 0.169, which recovers 63% of what a map fitted on the perturbed text would (0.116). So recalibration carries over partially, and the remainder is why a refit is worth repeating whenever the source of your inputs changes. The fine-tuned Maya-v1 shows the same pattern: under the same condition its accuracy is 75% and its macro-ECE goes from 0.056 on clean text to 0.202 (clean-text map 0.160, own map 0.091).

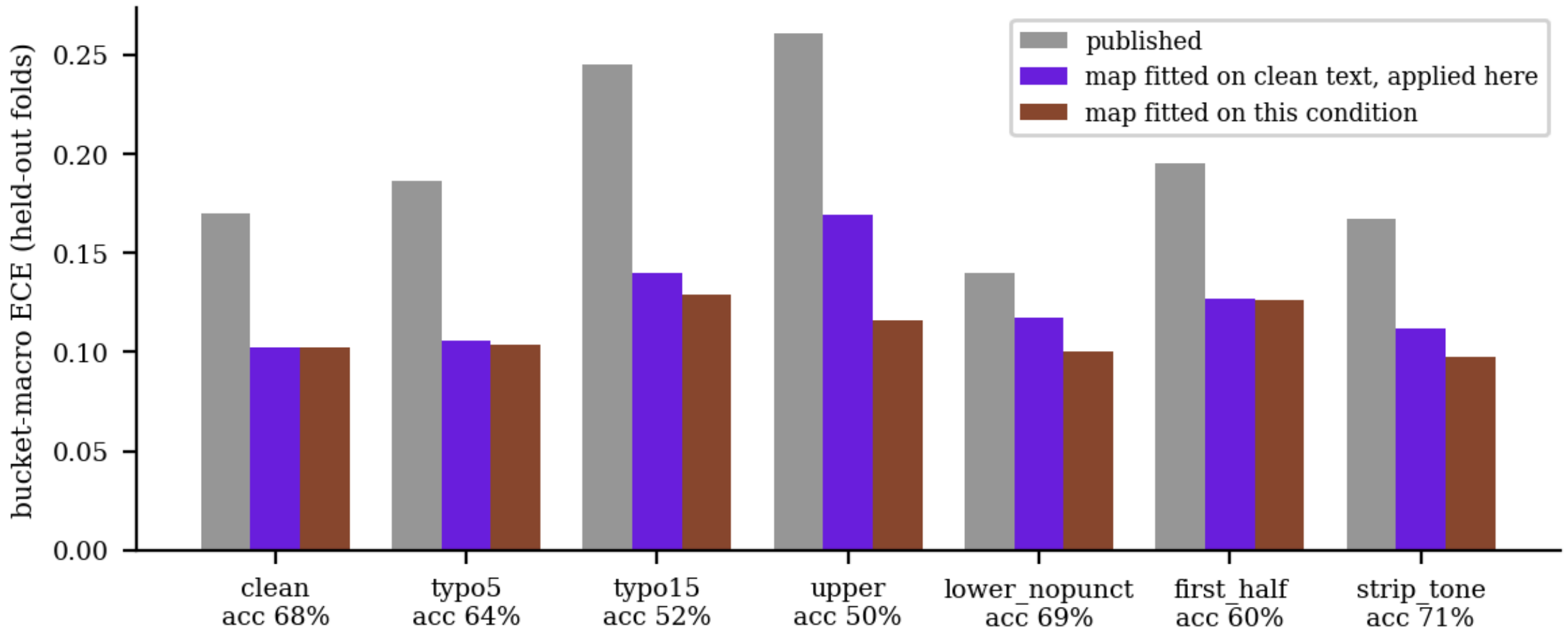

**Figure 20.** Pitfall 5: calibration depends on the input distribution. A histogram map is fitted on the clean text of four sets and then applied to the same items after perturbation (typos, upper case, truncation, tone removal). It recovers part of the error, and a map refitted on the perturbed text does better.

### 12.6 A short table of symptoms

**Table 13.** Symptoms, causes and cures at a glance.

| Symptom | Likely cause | What to do |
|---|---|---|
| A bucket's refit multiplier is ten times larger than any other | The stored temperature was clamped; the runtime is not applying what the file says | Compare stored and applied (Figure 19); fit on the published probabilities, so the multiplier already accounts for the clamp |
| Config has only three temperatures, all equal to 1 or 5 | As in the retrained checkpoints of this study, the saved config holds per-type values and no per-bucket table | Fit per-bucket multipliers yourself and write them back (Listing 5) |
| Per-bucket ECE jumps between runs | Bucket has a few dozen items | Report the chance floor; use temperature only; do not gate on that bucket until it has more items |
| Isotonic output is exactly 0 or 1 | Pooling made a block of items all wrong or all right | Clip to [0.02, 0.98] or switch to a temperature |
| Platt slope is in the hundreds or thousands | Plain Newton steps on a bucket whose confidences are nearly all 1 | Damp the step (`max_step` in Listing 3) and check the slope is of order 1 |
| Refit looks excellent, production does not | Items were scored with a map fitted on themselves, or the input distribution changed | Cross-fit (Listing 4); refit after any change of input source (Figure 20) |
| Gate accepts confidently wrong items after refit | A temperature cannot reorder items inside a bucket | Use a stricter cut-off for that bucket, or fine-tune to improve the ordering |
| Different ECE for the same model in two reports | Different bin counts or bin rules, or pooled versus bucket-macro | State the rule; compare with a chance floor (Figure 5) |

## 13 Does the recipe transfer to other checkpoints?

We repeated Steps 1 to 4 on 33 checkpoints, all evaluated on the same 1,065 items: the released English and multilingual models, a typed-decision variant, and 30 models fine-tuned with Laya's trainer on the generated data (seeds, data-ablation and augmentation variants). Table 14 shows the headline ones and Figure 21 shows all.

**Table 14.** Selected checkpoints: accuracy on the pool, mean confidence, pooled ECE, bucket-macro ECE before and after the cross-fitted refit.

| Checkpoint | Acc. | Mean conf. | Pooled ECE | Macro-ECE | Macro-ECE refit |
|---|---|---|---|---|---|
| Laya (root) | 69.4% | 0.721 | 0.073 | 0.143 | 0.073 |
| Laya typed-decisions | 70.5% | 0.636 | 0.124 | 0.181 | 0.078 |
| Laya-multilingual | 64.5% | 0.836 | 0.195 | 0.306 | 0.126 |
| Maya-v1 | 86.9% | 0.937 | 0.071 | 0.059 | 0.029 |
| Maya-multilingual | 91.5% | 0.930 | 0.027 | 0.076 | 0.060 |

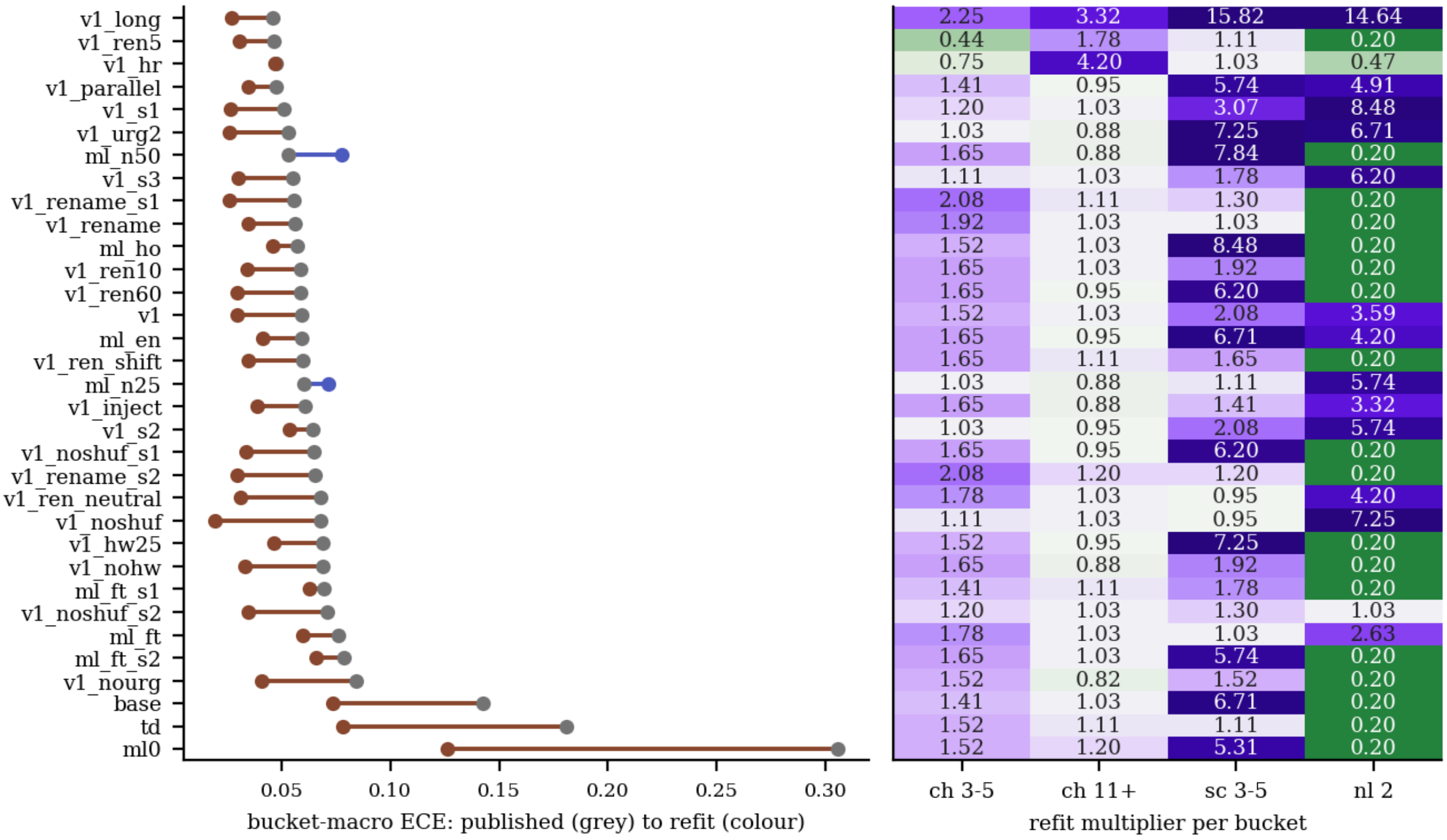

**Figure 21.** Step 3 repeated on 33 checkpoints that were evaluated on the same 1,065 items. Left: bucket-macro ECE before (grey) and after (green if better, red if worse) the cross-fitted per-bucket temperature, sorted by the published value. Right: the multiplier each bucket needs (purple: soften, orange: sharpen; blank: fewer than ten items; 0.20 is the bottom of the search grid). Only a few checkpoints, the released English one ('base') among them, need a 12-option multiplier above 1.5.

The cross-fitted per-bucket temperature lowers bucket-macro ECE for 31 of 33 checkpoints and does not for 2 (ml_n25, ml_n50). That is the first fold assignment; over 10 assignments 29 checkpoints improve every time and 4 do not (ml_ho in 9 of 10, ml_n25 in 5 of 10, ml_n50 in 8 of 10, v1_hr in 9 of 10). The recipe was designed on the base checkpoint, which is one of the 33, and every checkpoint is scored on the pool its fit comes from, so the count is an in-pool one: with a whole set held out the refit helps 21 of 33 (Section 8.4). A likely reason for the failures is that the published confidence was already close to what a refit can reach while the cross-fit noise at 10-681 items per bucket dominates, but we did not test this. The published macro-ECE of the base model (0.143) is among the highest in the set, and the multilingual root checkpoint is worse still (0.306). The retrained models mostly start low (median 0.059) and end lower (median 0.035). They are variants of one training recipe on the same generated rows, so they are not 30 independent samples.

On this pool, a large 12-option correction appears in only a few checkpoints: 30 of the 33 checkpoints with enough 12-option items need a multiplier between 0.8 and 1.25 there, and the ones above 1.5 are base, ml0, td. We did not test other data or other releases. Across the four seeds of the main retrained model (v1 and three reruns) the raw macro-ECE is 0.059, 0.051, 0.064, 0.055 and the refit 0.029, 0.026, 0.054, 0.030. The seed-to-seed standard deviation of the raw value is 0.005, so differences of that size between two checkpoints are not evidence of better calibration.

## 14 The checklist

**Table 15.** The checklist. Tick every row before a confidence threshold goes into a system. With time for half, do rows 1-4: they are cheap, and if every bucket is within its chance floor nothing needs fixing.

| # | Do | Pass when |
|---|---|---|
| 1 | Log the probability vector (or logits), the bucket and the reviewed label for each item | at least 20 labelled items in every bucket you will gate on, 40 or more if you want a confidence-only map |
| 2 | Compute pooled ECE and bucket-macro ECE with the same bins (Listing 1) | both reported, with a bootstrap interval; their ratio is 2.0 here |
| 3 | Compare each bucket's ECE with its chance floor (Figure 5) | buckets inside their floor are left alone; others have a sign (over or under) |
| 4 | Check what the runtime applies: stored versus clamped values | stored and applied temperatures are written next to each other (Table 2) |
| 5 | Fit a multiplier per bucket on published probabilities (Listing 2) | multiplier interval is narrower than the gap between over- and under-confident buckets |
| 6 | Cross-fit and repeat with several fold assignments (Listing 4) | macro-ECE after refit is stable to about 0.006 and below the published value |
| 7 | Re-run the binning and clean-label checks | ordering of methods and conclusions unchanged |
| 8 | Write the temperatures back and choose cut-offs on held-out items (Listings 5 and 6) | the gate's error is within the target in every bucket you accept from, or that bucket abstains |
| 9 | Look at the most confident mistakes (Appendix B) | you know which kinds of item still pass the gate |
| 10 | Schedule a refit | after any retraining, change of text source or template (shift test) |

## 15 Limitations

- The data are synthetic and the labels LLM-made. Accuracy here is agreement with the teacher, and calibration cannot correct a biased teacher [Sudjianto et al. 2026]. Real traffic will have its own label noise and bucket mix.
- The calibration fit and the evaluation share the same 1,065-item pool, kept apart only by five-fold cross-fitting: no item is scored by a map fitted on it, but fit and test items come from the same generator, labelling model and bucket mix, and no pool from another source or date is held out; holding out a whole set (Section 8.4) is still the same generator. The only check on changed inputs is the perturbed-text test in Section 12. A production recalibration should fit on one reviewed set and be judged on another, collected later.
- Buckets are an approximation: we assign items to Laya's type-by-option-count buckets, and urgency items are scored as `score:3-5` and churn items as `noul:2` by category. No item in the pool has two, six to ten options, so `choice:2` and `choice:6-10` are untested.
- Only the top-label probability is calibrated; the full vector and the joint behaviour of related questions, studied separately [Li et al. 2026; Joy 2026], are not covered.
- All ECE values use the binned estimator. Kumar et al. [2019] show that such estimates can understate the error of scaling methods, and a kernel-smoothed ECE [Błasiok et al. 2023] avoids the binning, but Table 8 varies the bins only. Dirichlet calibration [Kull et al. 2019] and scaling-binning [Kumar et al. 2019] were not tried, so "best method" means best of four on this pool.
- Strong simple baselines exist: a TF-IDF classifier matches or beats the fine-tuned Maya-v1 on 1 of 5 sets, and the fine-tuned models are fragile to perturbed text (Figure 20).

## 16 Conclusion

On the released Laya checkpoint the pooled ECE is 0.073 but the bucket-macro ECE is 0.143, because the 12-option bucket is over-confident by +0.22, the three-to-five-option bucket under-confident by 0.05, and the two cancel. Five steps and 130 lines of numpy remove about 49% of the macro-ECE on random folds and 31% when a whole set is held out, without touching the weights. A per-bucket temperature takes macro-ECE to 0.073 (0.078 ± 0.006 over 10 fold assignments) and reaches 0.089 with only 10 labelled items per bucket. A gate aimed at 10% error goes from answering 6.9% of items at 13.0% error to 46.3% at 10.5%; one cut-off per bucket on the published confidences does almost as well (45.4% at 10.3%), and neither reliably meets the target. Calibration cannot remove confidently wrong answers inside a bucket, and it should be repeated after any retraining or change of text source. The pool is synthetic and LLM-labelled, so these findings describe these checkpoints.

## Appendix A A complete driver script

The listings of Steps 1-5 chain together as follows. The script was executed when this paper was built, on the pooled data, and the output shown is what it printed (the cross-fitted macro-ECE is asserted to equal the one in Table 7).

```
nll, top = temper_tables(probs, gold)
conf = np.array([p.max() for p in probs])
ok = np.array([p.argmax() == g for p, g in zip(probs, gold)], float)
b = np.array(bucket)
D = dict(conf=conf, ok=ok, b=b, nll=nll, top=top)

print("pooled ECE %.3f   macro-ECE %.3f" % (ece(conf, ok), macro_ece(conf, ok, b)))            # step 1
for u in np.unique(b):                                                                          # step 2
    m = b == u
    print("%-11s n=%4d  gap %+.3f" % (u, m.sum(), conf[m].mean() - ok[m].mean()))
mult = {str(u): float(fit_temperature(nll[b == u])) for u in np.unique(b) if (b == u).sum() >= 10}   # step 3
print({u: round(v, 2) for u, v in mult.items()})
cal = crossfit(D, by_temperature)                                                               # step 4
print("macro-ECE after cross-fit %.3f" % macro_ece(cal, ok, b))
cfg = recalibrated_config(cfg, mult)                                                            # step 5
print({u: round(v, 2) for u, v in cfg["temperature_by_options"].items()})
```

Output.

```
pooled ECE 0.073   macro-ECE 0.143
choice:11+  n= 270  gap +0.225
choice:3-5  n= 681  gap -0.054
noul:2      n=  25  gap -0.117
score:3-5   n=  89  gap +0.083
{'choice:11+': 4.2, 'choice:3-5': 0.75, 'noul:2': 0.47, 'score:3-5': 1.03}
macro-ECE after cross-fit 0.073
{'choice:3-5': 1.33, 'choice:6-10': 1.0, 'score:3-5': 1.29, 'noul:2': 0.94, 'choice:11+': 2.1, 'choice:2': 1.91}
```

## Appendix B The most confident mistakes

Table B-1 lists the most confident mistakes in each bucket, so you can judge whether they share a pattern before you rely on a threshold.

**Table B-1.** The most confident mistakes of the base checkpoint (up to three per bucket; multilingual items excluded). Text is cut at 110 characters.

| Bucket | Set | Text | Predicted | Gold | Published conf. | Refit conf. |
|---|---|---|---|---|---|---|
| choice:11+ | many_test | Every time I try to enter my password, it spins forever and never lets me into my account dashboard. | password | login | 1.000 | 0.927 |
| choice:11+ | many_test | Who else has seen my bank transaction history? I see a log entry indicating a data export was performed on Mar… | security | privacy | 1.000 | 0.927 |
| choice:11+ | many_test | Please generate an RMA number so I can process the refund for the unsatisfactory software license key. | refund | returns | 1.000 | 0.927 |
| choice:3-5 | json_test | {"id": "TKT-910-001", "customer": "Meridian Logistics", "subject": "URGENT: Billing discrepancy on invoice #88… | billing | other | 0.952 | 0.985 |
| choice:3-5 | json_test | {"id": "TKT-9003", "customer": "Beta Test User 44", "subject": "Billing Question: Why is my invoice amount dif… | billing | technical | 0.937 | 0.978 |
| choice:3-5 | probe | Contrary to what the support bot suggested, this is not a billing discrepancy, but a feature request for batch… | billing | other | 0.929 | 0.973 |
| score:3-5 | urg_test | Legacy migration script failed at step 4 of 10. The transaction log shows a deadlock error on table 'user_sess… | blocking | soon | 0.837 | 0.829 |
| score:3-5 | probe | HEY!! i've been staring at this screen for 3 hours and it's driving me BANANAS!! can someone please just help … | blocking | soon | 0.769 | 0.759 |
| noul:2 | probe | I was a customer for ten years, but I've made up my mind that I'm leaving today. I just need the removal link. | 1 | 0 | 0.764 | 0.924 |

## Appendix C Provenance and statistics

Data. The pool is the union of 7 held-out sets (Table 4), 1,065 items, each item one question. Generation and filtering are described in Section 3; the independent model agrees with the stored label on 94%-99% of the items of each set.

Statistics. Bootstrap intervals [Efron and Tibshirani 1993] resample items with replacement (300 resamples for the ECE-type quantities and for the temperature intervals; 2.5th and 97.5th percentiles). Cross-fitting uses five folds from a seeded permutation; the repeats use 10, 20 and 8 permutations for the methods table, the temperature stability and the fit-set size. The chance floor simulates 300 label vectors per bucket, each item correct with probability equal to its own confidence.

Artefacts. Every number is computed at build time from the frozen per-item results (SHA-256 of the base results file begins `12eebd97c7af1a9b`).

## References

1. Laya: multilingual non-autoregressive decision model. *huggingface.co/convaiinnovations/laya*; code *github.com/NandhaKishorM/laya*.
2. Guo et al. On Calibration of Modern Neural Networks. arXiv:1706.04599, 2017.
3. Naeini, Cooper and Hauskrecht. Obtaining Well Calibrated Probabilities Using Bayesian Binning. *AAAI*, 2015.
4. Nixon et al. Measuring Calibration in Deep Learning. arXiv:1904.01685, 2019.
5. Kull et al. Beyond temperature scaling: Obtaining well-calibrated multi-class probabilities with Dirichlet calibration. arXiv:1910.12656, 2019.
6. Ovadia et al. Can You Trust Your Model's Uncertainty? Evaluating Predictive Uncertainty Under Dataset Shift. arXiv:1906.02530, 2019.
7. Desai and Durrett. Calibration of Pre-trained Transformers. arXiv:2003.07892, 2020.
8. Geifman and El-Yaniv. Selective Classification for Deep Neural Networks. arXiv:1705.08500, 2017.
9. Brier. Verification of forecasts expressed in terms of probability. *Monthly Weather Review* 78, 1950.
10. Warner et al. Smarter, Better, Faster, Longer: ModernBERT. arXiv:2412.13663, 2024.
11. Platt. Probabilistic outputs for support vector machines and comparisons to regularized likelihood methods. In *Advances in Large Margin Classifiers*, 1999.
12. Zadrozny and Elkan. Transforming classifier scores into accurate multiclass probability estimates. *KDD*, 2002.
13. Niculescu-Mizil and Caruana. Predicting good probabilities with supervised learning. *ICML*, 2005.
14. Chow. On optimum recognition error and reject tradeoff. *IEEE Transactions on Information Theory* 16(1), 1970.
15. Kumar, Liang and Ma. Verified Uncertainty Calibration. *NeurIPS*, 2019. arXiv:1909.10155.
16. Minderer et al. Revisiting the Calibration of Modern Neural Networks. *NeurIPS*, 2021. arXiv:2106.07998.
17. Vaicenavicius et al. Evaluating model calibration in classification. *AISTATS*, 2019. arXiv:1902.06977.
18. DeGroot and Fienberg. The comparison and evaluation of forecasters. *The Statistician* 32, 1983.
19. Błasiok et al. Smooth ECE: Principled Reliability Diagrams via Kernel Smoothing. arXiv:2309.12236, 2023.
20. Berta et al. CalArena: A Large-Scale Post-Hoc Calibration Benchmark. arXiv:2605.30188, 2026 (preprint).
21. Sharma, Duncan, Marceau Caron and Vollmer. TypedBench: A Benchmark for Calibration, Framing Sensitivity, and Cost in System One Decision Models. arXiv:2610.11392, 2026 (preprint).
22. Teke. ufakzeka-karar: An Open Turkish Typed-Decision Model with Order-Invariant Option Scoring. arXiv:2610.06744, 2026 (preprint).
23. Li, He and Li. Beyond Calibration: Do a Typed-Decision Model's Probabilities Obey the Probability Axioms? arXiv:2609.33209, 2026 (preprint).
24. Joy. Do System One Decisions Add Up? A Study of Probabilistic Coherence. arXiv:2609.33971, 2026 (preprint).
25. Rafe and Das. Benchmarking System One decision models against trained classifiers and language models for automated decision gates. arXiv:2610.00346, 2026 (preprint).
26. Porcedda. Jev thinks "I don't know", but doesn't say it: Introducing Sys1Cal-v1 Dataset for Probability Calibration. arXiv:2609.35342, 2026 (preprint).
27. Dai et al. ProbabilityBench: Model Evaluation and Calibration with Exact Ground-Truth Uncertainty. arXiv:2610.17016, 2026 (preprint).
28. Santillana. VX-One: Typed, Calibrated Decision Heads for Real-Time Perimeter Triage. arXiv:2610.30090, 2026 (preprint).
29. Sudjianto et al. Teacher Supervision, Fixed Representations and Calibration: Exact Risk Calculations for Candidate-Scoring Decision Models. arXiv:2610.26797, 2026 (preprint).
30. Efron and Tibshirani. *An Introduction to the Bootstrap*. Chapman & Hall, 1993.